\documentclass[letterpaper, 10 pt, conference]{ieeeconf}  % Comment this line out if you need a4paper

\IEEEoverridecommandlockouts                              % This command is only needed if 
\usepackage{graphicx} % for pdf, bitmapped graphics files
\usepackage{amsmath} % assumes amsmath package installed
\usepackage{amssymb}  % assumes amsmath package installed
\usepackage{subcaption}

\title{\LARGE \bf
An Insight on the Ratio of Transmission of Motion (RoToM)\\and its Relation to the Centroidal Inertia Matrix
}

\author{Federico L. Moro% <-this % stops a space
\thanks{Federico L. Moro is with the Institute of Industrial Technologies and Automation (ITIA), National Research Council (CNR) of Italy, Via Corti 12, 20133 Milano, Italy
        {\tt\small federico.moro@itia.cnr.it}}%
}

\begin{document}

\maketitle
\thispagestyle{empty}
\pagestyle{empty}

%%%%%%%%%%%%%%%%%%%%%%%%%%%%%%%%%%%%%%%%%%%%%%%%%%%%%%%%%%%%%%%%%%%%%%%%%%%%%%%%
\begin{abstract}
This paper analyses the dynamic response of a robot when subject to an external force that is applied to its Center of Mass (CoM). The \emph{Ratio of Transmission of Motion (RoToM)} is proposed as a novel indicator of what part of the applied force generates motion, and what part is dissipated by the passive forces due to mechanical constraints. It depends on the configuration of the robot and on the direction of the force, and is always between 0 and 1. Extending this concept, a \emph{transmissibility ellipsoid} is used to describe the behavior of the robot given a certain configuration, and varying the direction of the applied force. Another physical measure that is related to the transmissibility ellipsoid is the \emph{transmissibility index}: it provides an indication on how similarly the system behaves when subject to forces coming from different directions.
The presented analysis aims to provide a deeper insight on the centroidal dynamics of a robot, and on its dependence on the configuration. It can be beneficial for developing whole-body controllers of redundant robots for e.g., reducing the effort in terms of joint torques to compensate for gravity, and more in general for designing interaction control architectures.
\end{abstract}

%%%%%%%%%%%%%%%%%%%%%%%%%%%%%%%%%%%%%%%%%%%%%%%%%%%%%%%%%%%%%%%%%%%%%%%%%%%%%%%%
\section{INTRODUCTION}
The first industrial robots have entered factories in early 1960's, and in the last half-century their use has revolutionized manufacturing. Recent developments in robotics research now open the doors to a brand new class of robots that are going to change people's everyday life. These robots co-operate side-by-side with humans in real world scenarios. For this reason, interaction with humans and with the environment plays a key role, and needs to be studied and controlled accordingly.

Interaction happens when a force is exchanged between the robot and the person/object that is in contact with it, and is a well studied paradigm. Excellent surveys on this topic can be found in the literature \cite{Siciliano}, \cite{Hogan}, \cite{DeSantis}.

Another force of different nature that affects the dynamics of a robot is gravity, that can be modeled as a constant force that is applied to the Center of Mass (CoM) of the robot. The dynamic response of a robot to an external force that is applied depends significantly on its configuration (i.e., its joint angles). A redundant robot, for instance, can assume different configurations depending on the desired behavior.

Based on this consideration, the Reaction Mass Pendulum (RMP) model \cite{Lee} was proposed to describe the dynamic behavior of robots, particularly bipeds. Differently from other reduced-dimensionality models it includes the information on the centroidal inertia \cite{Orin}. This feature is particularly important for momentum-based controllers \cite{Moro1}, \cite{Koolen}, \cite{Wensing}.

This paper investigates the dynamics of a robot when subject to a force. In Section II, it introduces the \emph{Ratio of Transmission of Motion (RoToM)}, a novel indicator of what part of the force that is applied to the CoM of a robot generates motion, and what part is dissipated by passive forces due to physical constraints. The \emph{RoToM} depends on the configuration of the robot, and on the direction of the applied force. A simple example with a single pendulum moving on a plane is presented in Section III. Other two physical concepts that are related to the \emph{RoToM} are the \emph{transmissibility ellipsoid} and the \emph{transmissibility index}, and are described in Section IV. Section V discusses a possible generalization of what introduced in the previous sections, and Section VI presents the conclusions of the paper.

\begin{figure}[!t]
\centering
\includegraphics[width=0.9\linewidth]{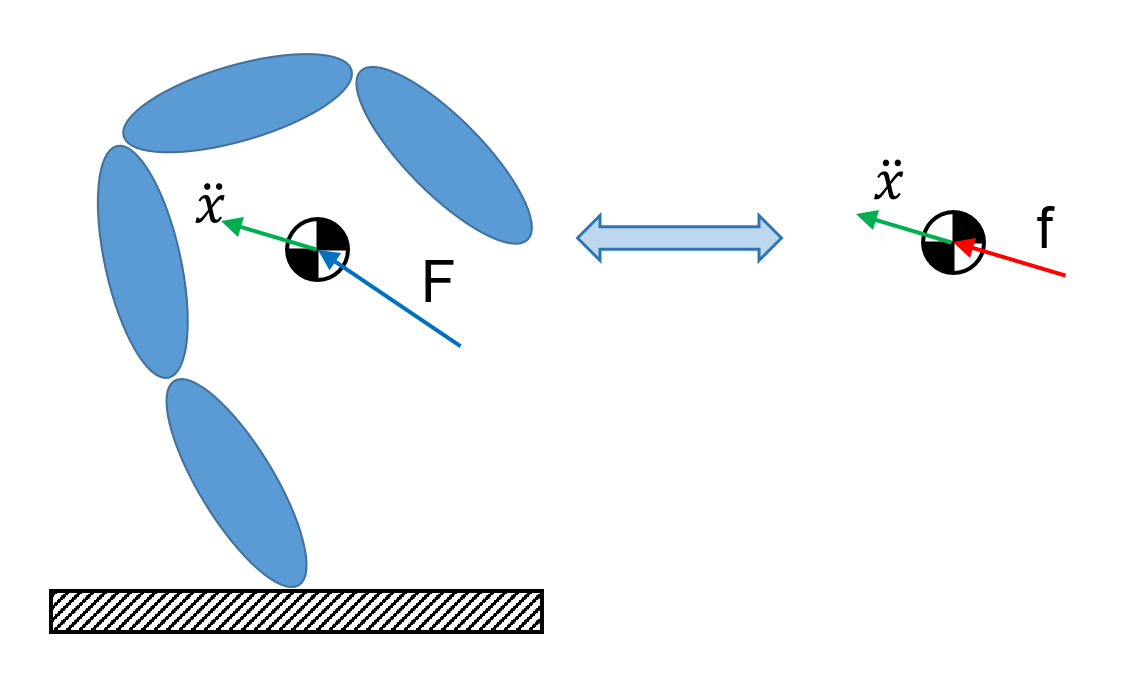}
\caption{A generic robot with mass $m$ and its corresponding representation as a point mass. $f(q)$ is a fictitious force that, if applied to an unconstrained point mass with mass $m$, produces a dynamic behavior that is the same as the one of the robot when a force $F$ is applied to its CoM.}\label{fig:model}
\end{figure}

\section{THE RATIO OF TRANSMISSION OF MOTION}
The dynamics of an unconstrained point mass with mass $m$, subject to a linear force $F$ is described by Newton's Second Law:
\begin{equation}
\ddot x = \frac{1}{m} F
\end{equation}

In the case of a more complex system (e.g., a $n$-links pendulum, a serial manipulator, or any robot) a number of constraints have to be taken into account, and the dynamics is described by the following equation expressed in joint coordinates:

\begin{equation}
M \ddot q + C \dot q + h = \tau + J^T F
\end{equation}

\noindent where $M(q)$, $C(q,\dot q)$, $h(q)$ are the inertia matrix, the Coriolis term, and the gravity vector of the system, respectively. $\tau$ is the vector of joint torques, and $J(q)$ is the Jacobian to the point where the force $F$ is applied.

In static conditions (i.e., $\dot q = 0$), with no joint torque $\tau$ applied, assuming that $F$ is a linear force that is applied to the Center of Mass (CoM) of the system, and considering gravity as an external force (i.e., $h = -J_c^T mg$)\footnotemark[1], Eq. 2 can be simplified as:

\footnotetext[1]{If $J_{c,i}$ is the Jacobian to the $i$-th link CoM, and $m_i$ is its mass, then the Jacobian to the robot CoM is defined as: $J_c=\frac{\sum_i \left( m_i J_{c,i} \right)}{\sum_i m_i}$}
 
\begin{equation}
M \ddot q = J^T F
\end{equation}

Eq. 3 can also be written in Cartesian coordinates with respect to the point where the force $F$ is applied (i.e., CoM):

\begin{equation}
\Lambda \ddot x = J^{-T}M J^{-1} \ddot x = F
\end{equation}

\noindent where $\Lambda(q)$ is the centroidal inertia matrix \cite{Lee,Orin}.

Notice that, in the general case, $J$ is not a squared matrix. The notation $J^{-1}$ is used for simplicity, while it actually is a pseudo-inverse $J^\#$. The discussion on what pseudo-inverse is better to use is out of the scope of this paper. Several pseudo-inverses are often used in the literature \cite{Sentis,Mansard,Gienger}. A survey on such methods is given in \cite{MoroSentis}.

The centroidal acceleration $\ddot x$ given a force $F$ that is applied to the CoM, therefore, depends on the Jacobian to the CoM, and on the inertia $M$ of the system. Both these quantities depend on the configuration $q$ of the system.

\begin{equation}
\ddot x = J M^{-1}J^T F
\end{equation}

It can be noticed that Eq. 5 can be written in a form that is similar to the one of Eq. 1:

\begin{equation}
\ddot x = \frac{1}{m} \left( m J M^{-1} J^T \right) F
\end{equation}

This is easier to visualize if $f$ is defined as:

\begin{equation}
f = \left( m J M^{-1} J^T \right) F
\end{equation}

Combining Eqs. 6 and 7 the dynamics of the system is described in the same form as in Eq. 1:

\begin{equation}
\ddot x = \frac{1}{m} f
\end{equation}

$f(q)$ is a fictitious force that, if applied to an unconstrained point mass with mass $m$, produces a dynamic behavior that is the same as the one of the constrained system when a force $F$ is applied (Figure \ref{fig:model}). It is the vector addition of the force $F$, and the passive reaction forces $R$ due to the mechanical constraints of the system.

\begin{equation}
\vec{f} = \vec{F} + \vec{R}
\end{equation}

Given the nature of the passive forces $R$, the magnitude of vector $f$ is at most equal to the magnitute of vector $F$.

\begin{equation}
0 \le \frac{|| f ||}{|| F ||} \le 1
\end{equation}

The ratio between the magnitude of $f$ and the magnitude of $F$ is therefore always between $0$ and $1$. According to the definition of $f$ in Eq. 7, the ratio in Eq. 10 can be expanded to:

\begin{equation}
\frac{|| f ||}{|| F ||} = \frac{||m J M^{-1}J^T F||}{||F||}
\end{equation}

\noindent and, equivalently, to:

\begin{equation}
\frac{|| f ||}{|| F ||} = \left\lvert\left\lvert m J M^{-1}J^T \frac{F}{||F||}\right\rvert\right\rvert
\end{equation}

This quantity is a scalar between $0$ and $1$ that depends on the configuration of the robot $q$ and on the direction of vector $F$, and describes in what part the force applied generates motion, and is therefore named {\it Ratio of Transmission of Motion (RoToM)}. When the {\it RoToM} is equal to $1$ no part of the force $F$ is dissipated by the passive forces $R$. On the contrary, when it is equal to $0$ no motion is generated by the force $F$. The closer to $1$ the {\it RoToM} is, the more efficient the transmission of the applied force $F$ is.

\begin{equation}
0 \le \left\lvert\left\lvert m J M^{-1}J^T \frac{F}{||F||}\right\rvert\right\rvert = \left\lvert\left\lvert m \Lambda^{-1} \frac{F}{||F||}\right\rvert\right\rvert \le 1
\end{equation}

Notice that the {\it RoToM} is independent of both mass $m$ and force magnitude $||F||$. Eq. 13 can also be written from the perspective of the Cartesian accelerations. In a constrained system with mass $m$ and equivalent centroidal inertia $\Lambda$ the magnitude of the acceleration due to a linear force $F$ is always, as obvious, at most equal to the acceleration generated by the same force $F$ when applied to an unconstrained point mass with mass $m$.

\begin{equation}
0 \le \left\lvert\left\lvert \Lambda^{-1} F\right\rvert\right\rvert \le \frac{||F||}{m}
\end{equation}\\

\subsection{Local minimization of the RoToM}
The {\it Ratio of Transmission of Motion (RoToM)} is an indicator of what part of an applied force is transmitted to generate motion, and what part is dissipated by passive forces. Depending on the application it might be beneficial to have a high {\it RoToM} or a low {\it RoToM}.

In the case, for instance, of a robot subject to gravity, a smaller {\it RoToM} requires a lesser effort in terms of joint torques to compensate for it. Similarly to what presented in \cite{Moro1}, where an attractor to the {\it minimum effort} configuration is used together with a control on the {\it joint momentum} to guarantee balance in a whole-body motion control system, and extended in \cite{Moro2,Moro3} introducing the {\it gravitational stiffness}, a gradient descent of the {\it RoToM} (notice that $m$ and $||F||$ are both positive scalars and can be omitted in Eq. 15) represents a local minimization of the effort due to the external linear force $F$ (in the example, due to gravity) that is applied:

\begin{equation}
\tau_i = - k \frac{\partial \left\lvert\left\lvert \Lambda^{-1}(q)F \right\rvert\right\rvert}{\partial  q_i}
\end{equation}

Applying a vector of joint torques $\tau$ as in Eq. 15 will make the robot move towards a configuration such that a greater part of the external force $F$ is dissipated by the passive forces due to the mechanical constraints.

\subsection{Zeroes of the RoToM}
The method described in Section II.A is a local minimization, and can be very useful to solve practical problems. However, redundant robots will typically have several {\it RoToM} minima. Being able to identify all the configurations with zero {\it RoToM} can be very beneficial. Some of these, for instance, could be not reachable due to the robot's joint limits, making them not a valid solution to the problem of minimization.

Given the centroidal inertia matrix $\Lambda(q)$ of a robot, and a linear force vector $F$ that is applied to its CoM, the {\it RoToM} will be equal to zero if and only if $F$ is in the null-space of $\Lambda^{-1}$, i.e., find $q$ s.t. $F \in Ker(\Lambda^{-1}(q))$.

The solutions to this search problem are identified by solving the following system of equations:

\begin{equation}
\begin{cases}
\Lambda_{11}^{-1}(q) F_1 + \Lambda_{12}^{-1}(q) F_2 + \Lambda_{13}^{-1}(q) F_3 = 0\\
\Lambda_{21}^{-1}(q) F_1 + \Lambda_{22}^{-1}(q) F_2 + \Lambda_{23}^{-1}(q) F_3 = 0\\
\Lambda_{31}^{-1}(q) F_1 + \Lambda_{32}^{-1}(q) F_2 + \Lambda_{33}^{-1}(q) F_3 = 0\\
\end{cases}
\end{equation}\\

\section{A SIMPLE EXAMPLE}
\begin{figure}[!t]
\centering
\includegraphics[width=0.9\linewidth]{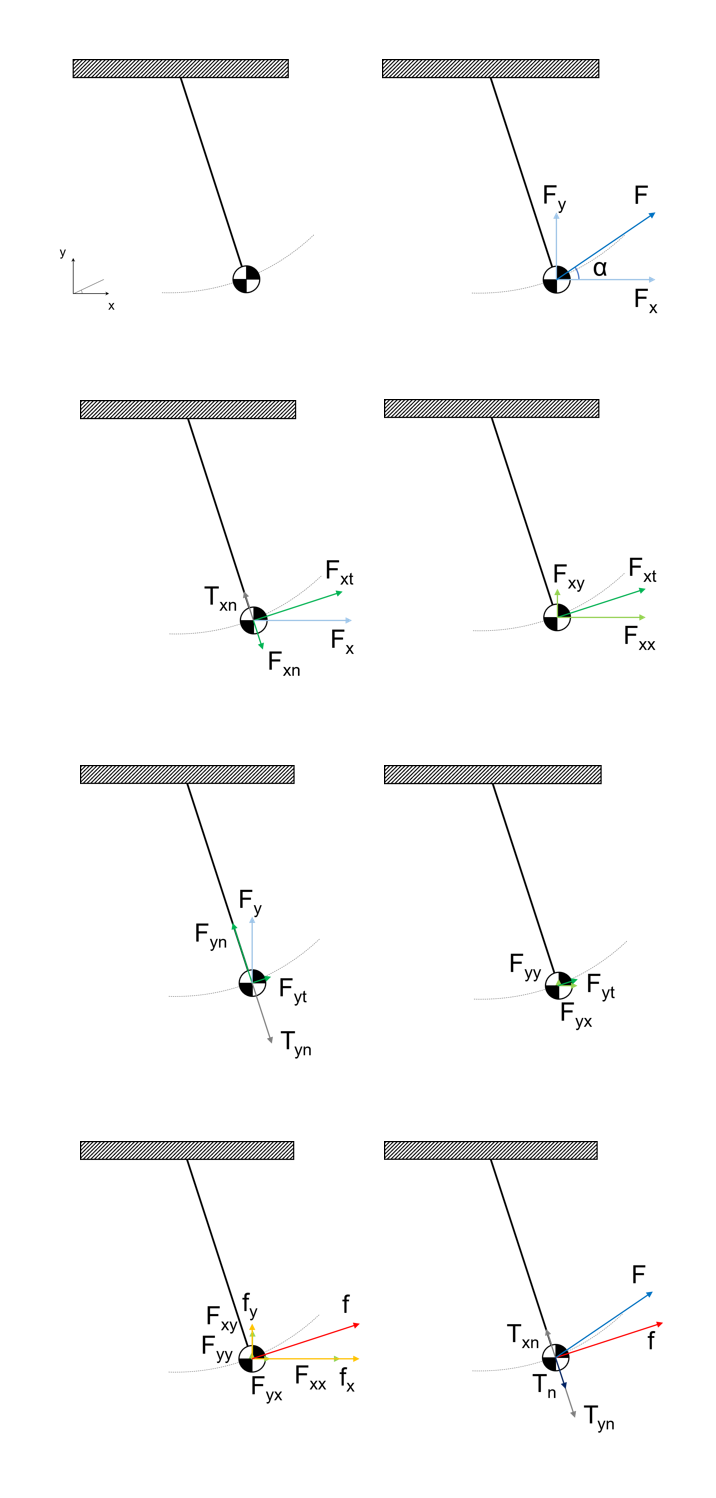}
\caption{Notation and conventions for the single pendulum example}\label{fig:forces_pen}
\end{figure}

The simplest example of constrained system is a single pendulum moving on a plane with a point mass with mass $m$ hanging at the end of a massless, rigid rod. Notation and conventions for this example are as in Figure \ref{fig:forces_pen}. A linear force $F$ is applied to the mass:

\begin{equation}
\begin{cases}
F_x = F \cos(\alpha)\\
F_y = F \sin(\alpha)
\end{cases}
\end{equation}

\noindent where $F_x$ and $F_y$ are the projections of the force $F$ onto the axes $x$ and $y$, respectively.

The force projection $F_x$ has two components: $F_{xn}$, which is compensated by $T_{xn}$, a passive force due to the constraint of the rod being rigid, and $F_{xt}$ that generates motion:

\begin{equation}
F_{xt} = - F_x \sin(q)
\end{equation}

\begin{equation}
\begin{cases}
f_{xx} = - F_{xt} \sin(q) = F_x \sin(q) \sin(q)\\
f_{yx} = F_{xt} \cos(q) = - F_x \sin(q) \cos(q)
\end{cases}
\end{equation}

\noindent where $f_{xx}$ and $f_{yx}$ are the projections of $F_{xt}$ onto the axes $x$ and $y$, respectively.

Similarly, $F_y$ has two components: $F_{yn}$, which is compensated by $T_{yn}$, a passive force due to the constraint of the rod being rigid, and $F_{yt}$ that generates motion:

\begin{equation}
F_{yt} = F_y \cos(q)
\end{equation}

\begin{equation}
\begin{cases}
f_{xy} = - F_{yt} \sin(q) = - F_y \sin(q) \cos(q)\\
f_{yy} = F_{yt} \cos(q) = F_y \cos(q) \cos(q)
\end{cases}
\end{equation}

\noindent where $f_{xy}$ and $f_{yy}$ are the projections of $F_{yt}$ onto the axes $x$ and $y$, respectively.

\begin{figure*}[!t]
\vspace{0.5cm}
\centering
\begin{subfigure}{.115\textwidth}
  \centering
  \includegraphics[width=\linewidth]{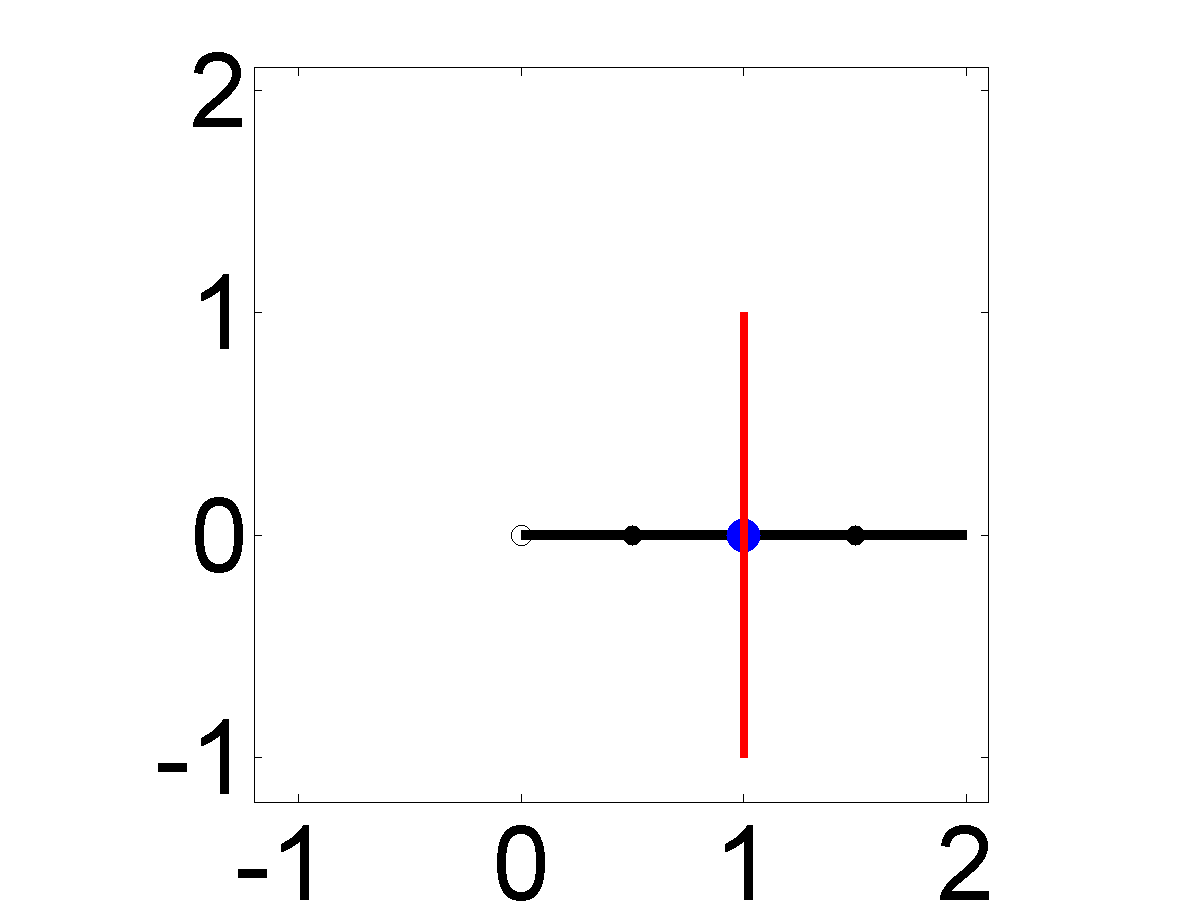}
\end{subfigure}
\begin{subfigure}{.115\textwidth}
  \centering
  \includegraphics[width=\linewidth]{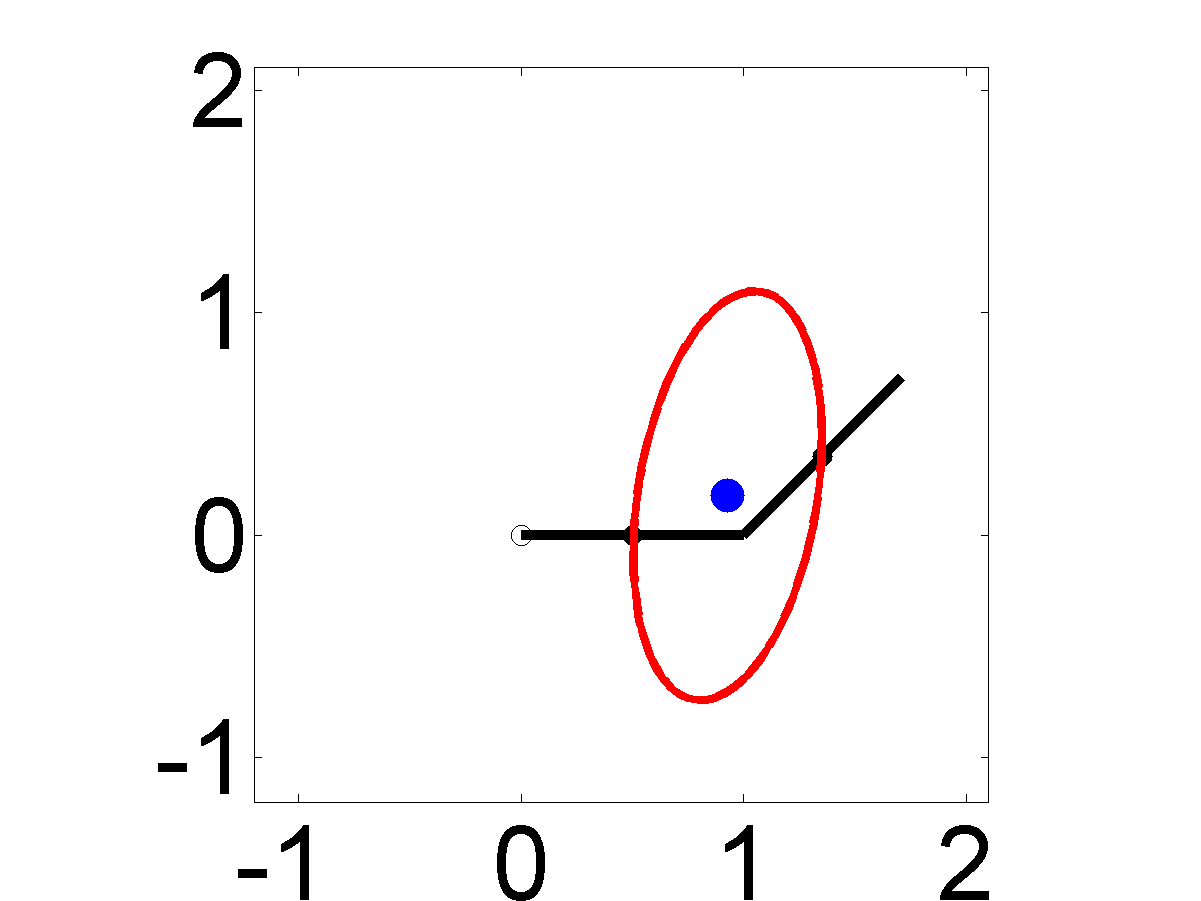}
\end{subfigure}
\begin{subfigure}{.115\textwidth}
  \centering
  \includegraphics[width=\linewidth]{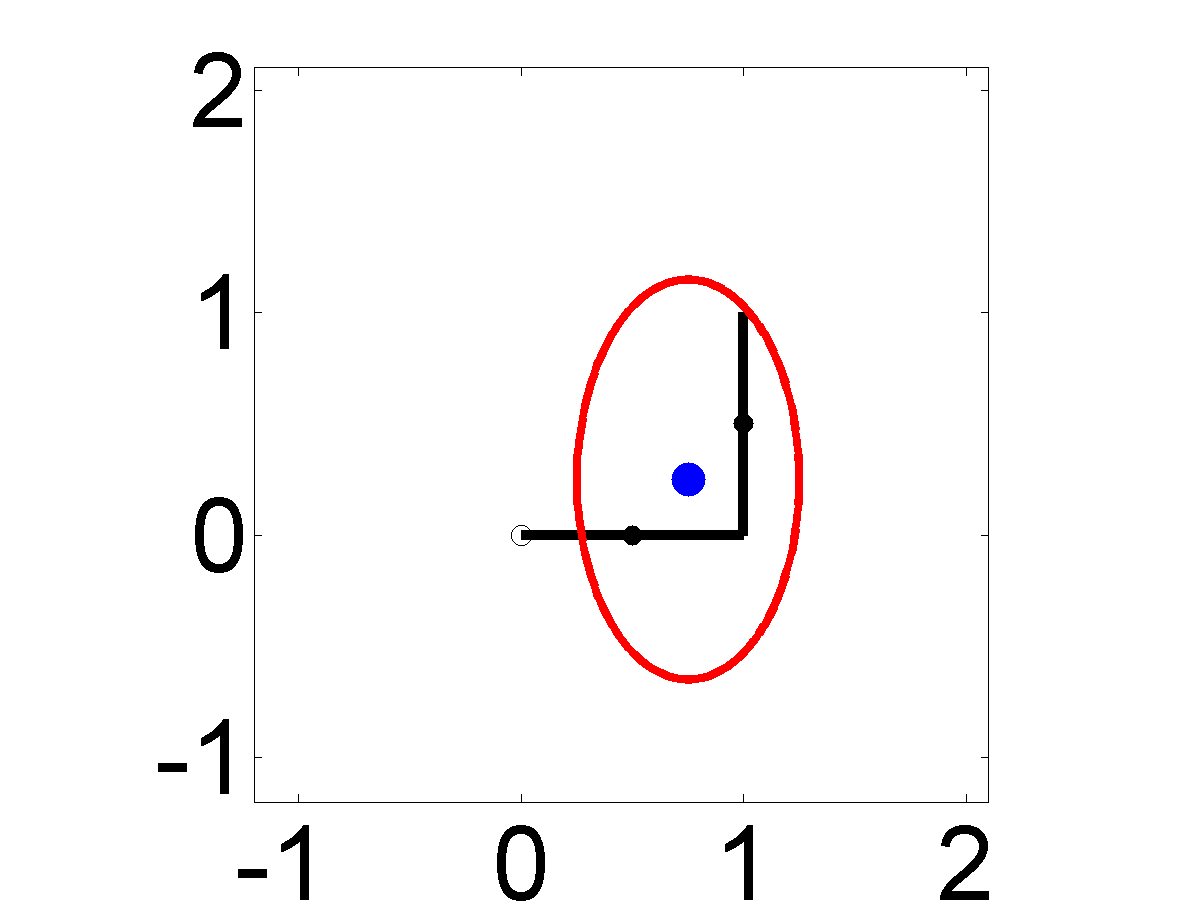}
\end{subfigure}
\begin{subfigure}{.115\textwidth}
  \centering
  \includegraphics[width=\linewidth]{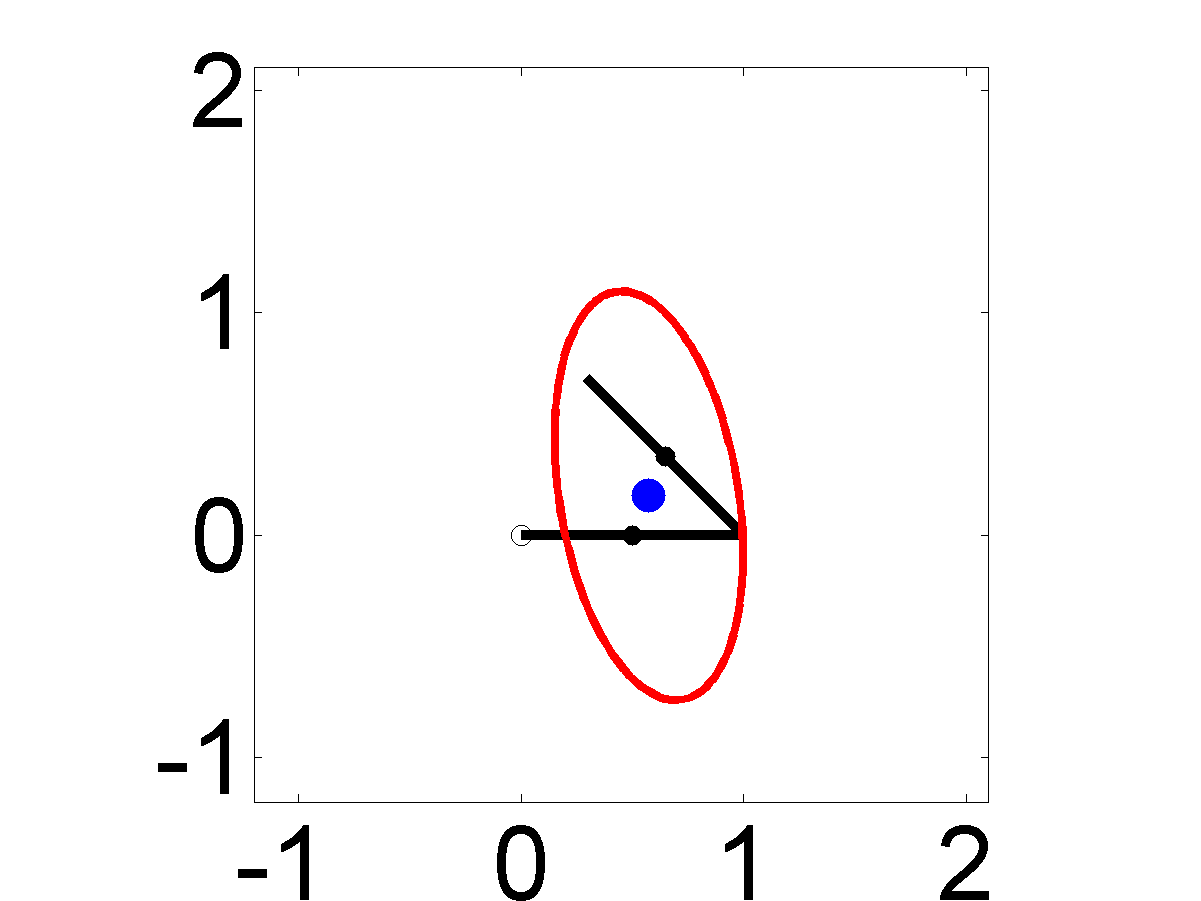}
\end{subfigure}
\begin{subfigure}{.115\textwidth}
  \centering
  \includegraphics[width=\linewidth]{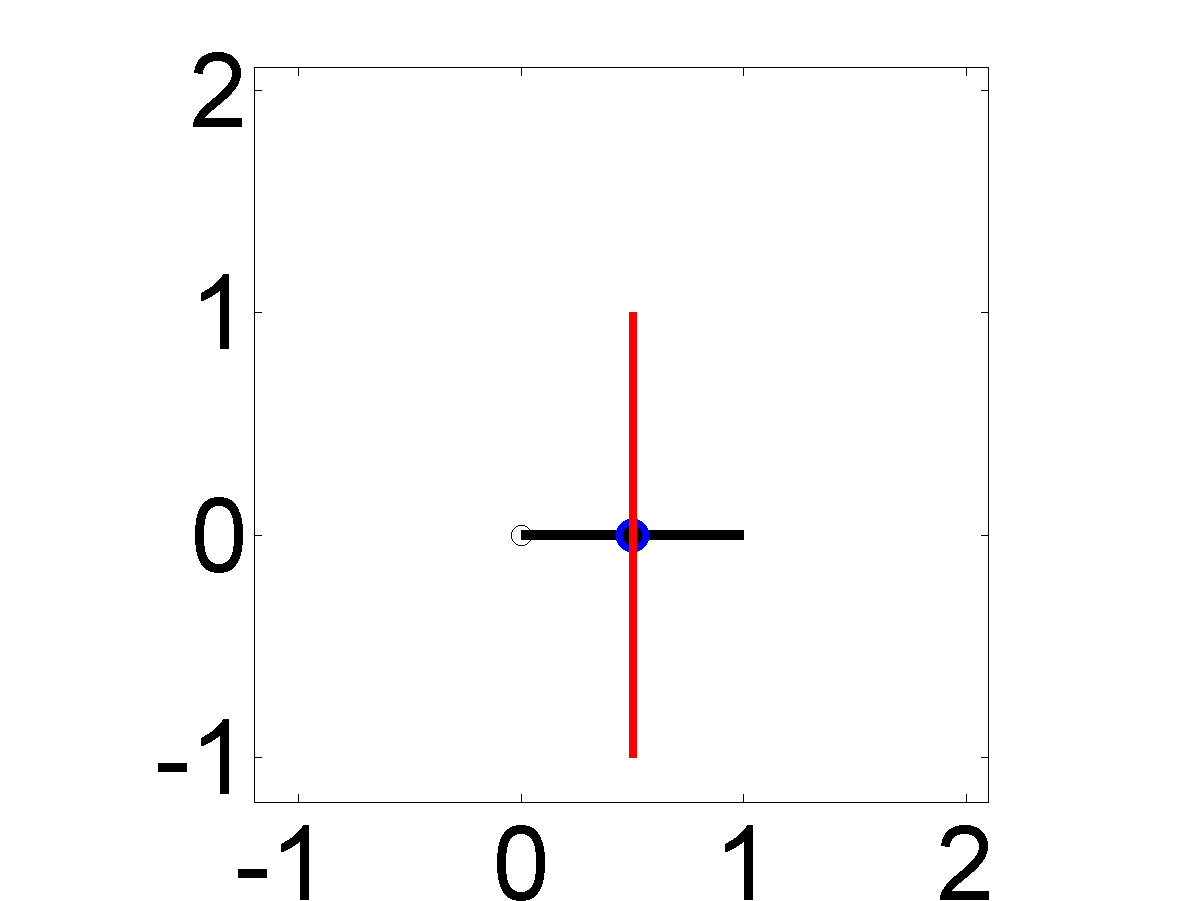}
\end{subfigure}
\begin{subfigure}{.115\textwidth}
  \centering
  \includegraphics[width=\linewidth]{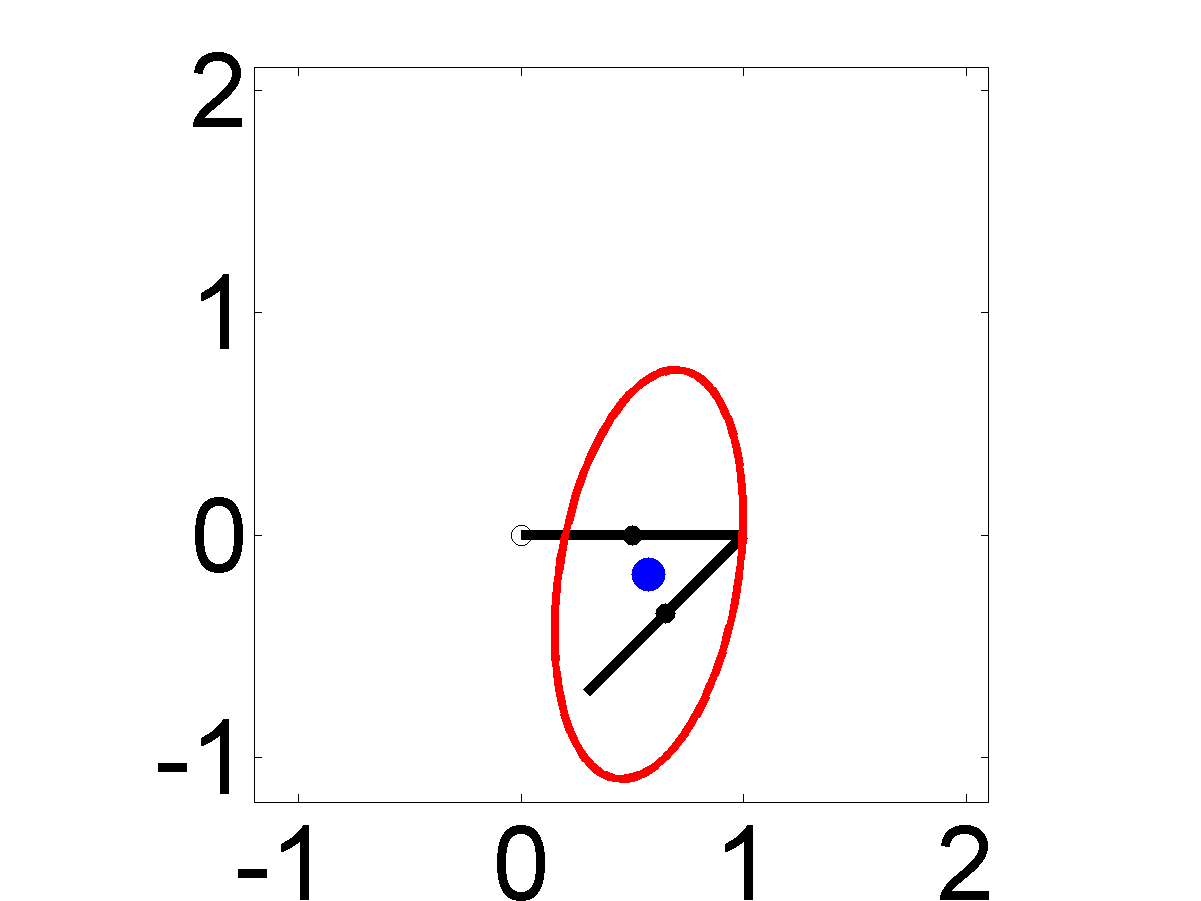}
\end{subfigure}
\begin{subfigure}{.115\textwidth}
  \centering
  \includegraphics[width=\linewidth]{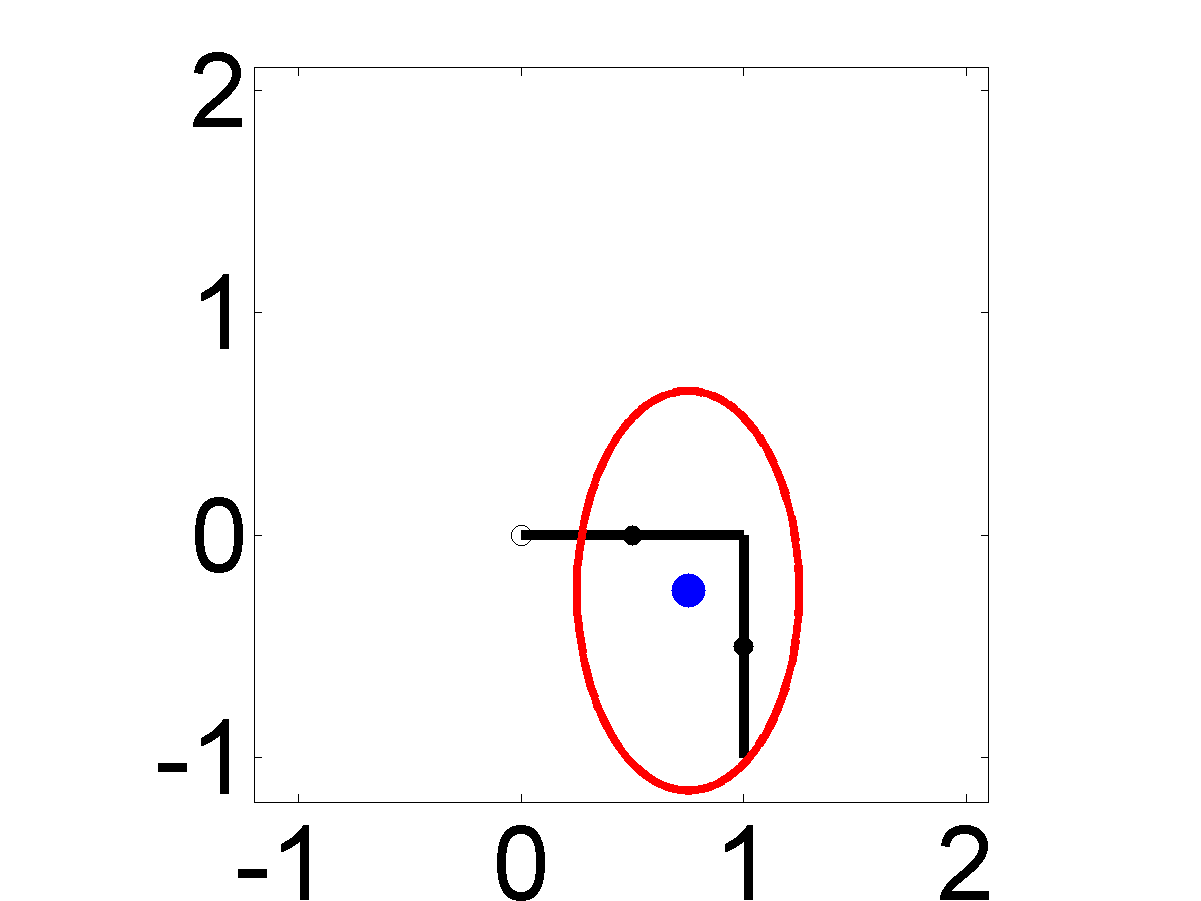}
\end{subfigure}
\begin{subfigure}{.115\textwidth}
  \centering
  \includegraphics[width=\linewidth]{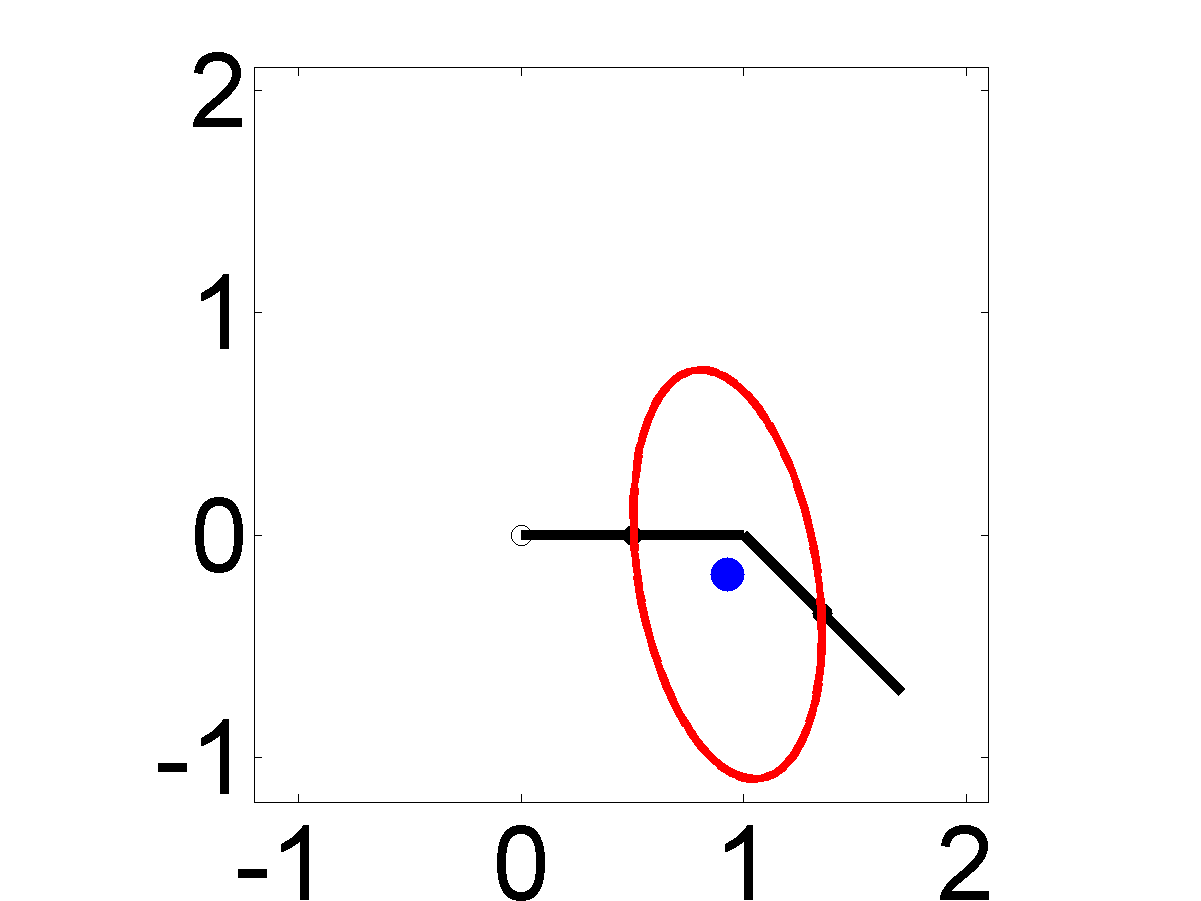}
\end{subfigure}
\begin{subfigure}{.115\textwidth}
  \centering
  \includegraphics[width=\linewidth]{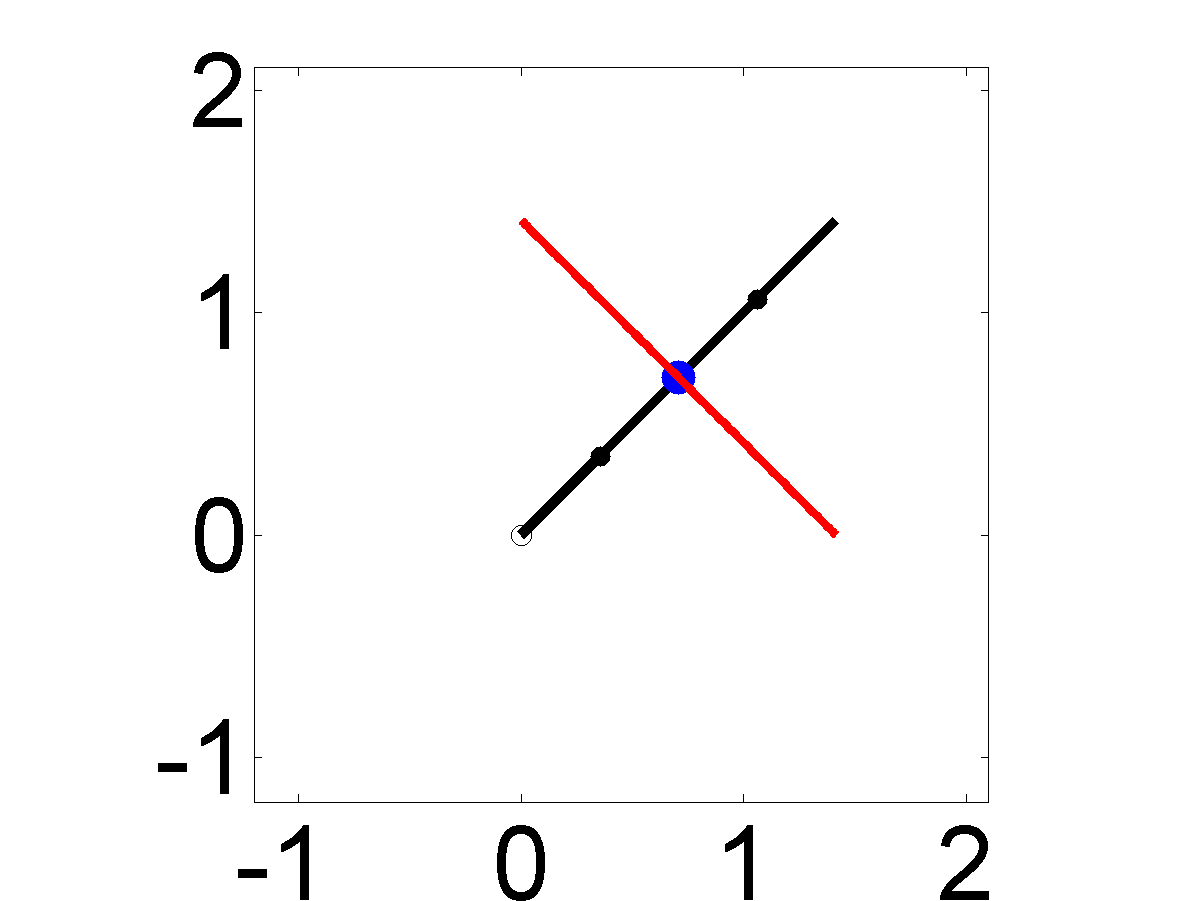}
\end{subfigure}
\begin{subfigure}{.115\textwidth}
  \centering
  \includegraphics[width=\linewidth]{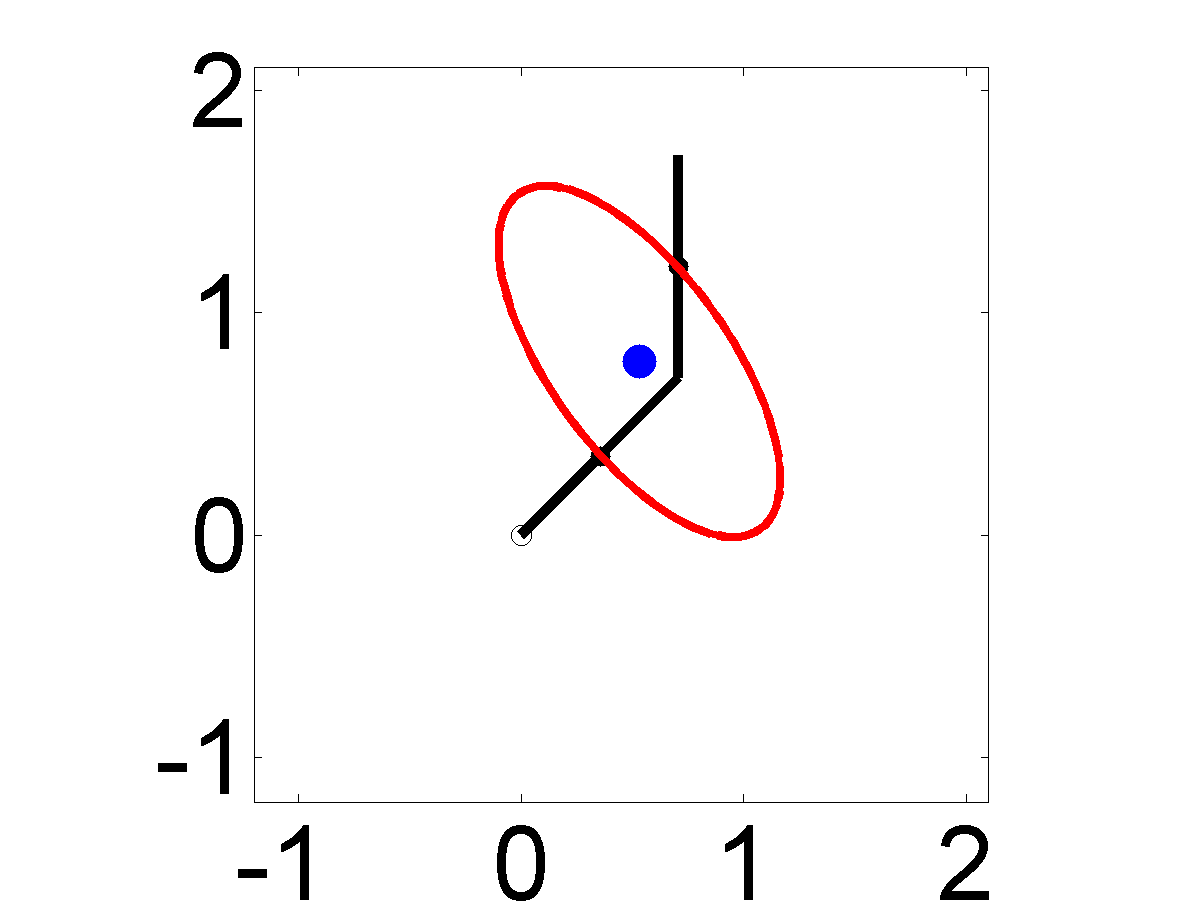}
\end{subfigure}
\begin{subfigure}{.115\textwidth}
  \centering
  \includegraphics[width=\linewidth]{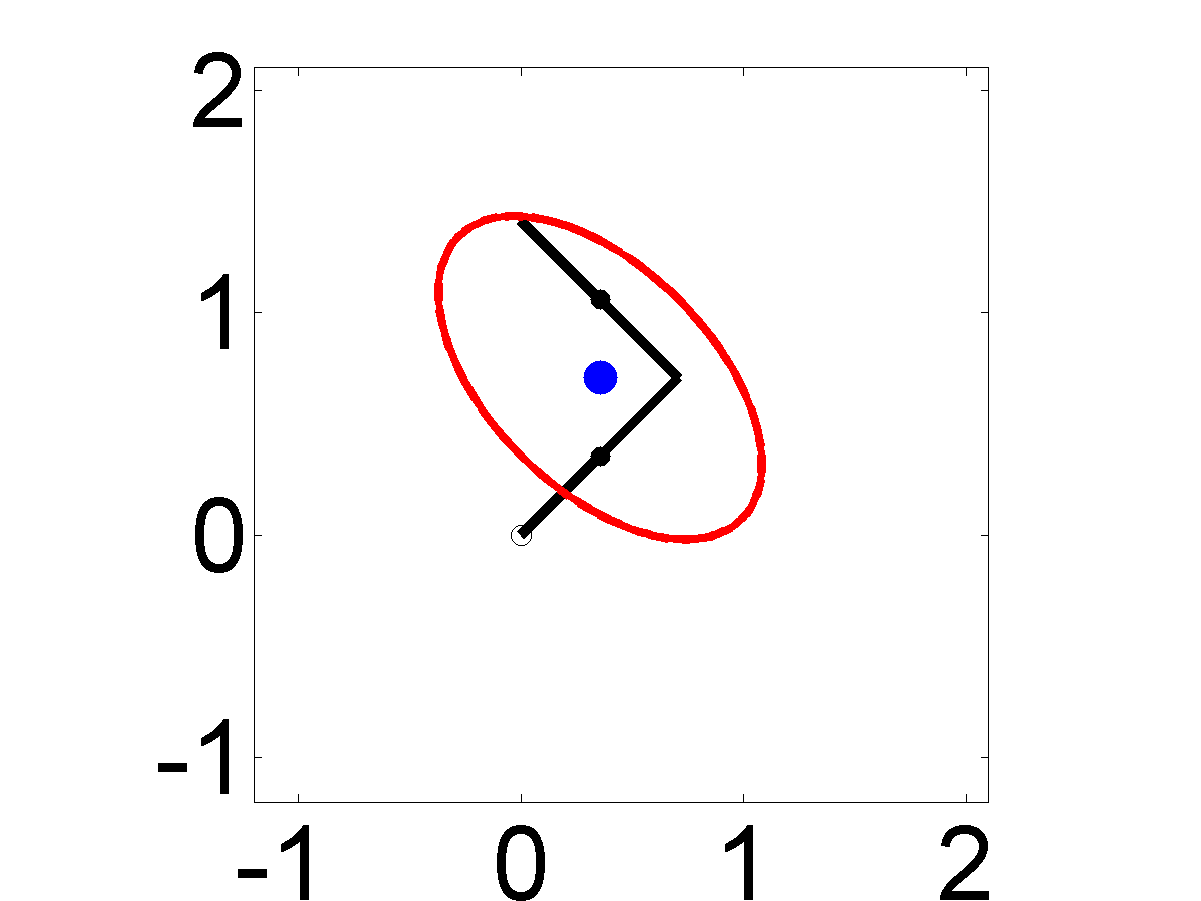}
\end{subfigure}
\begin{subfigure}{.115\textwidth}
  \centering
  \includegraphics[width=\linewidth]{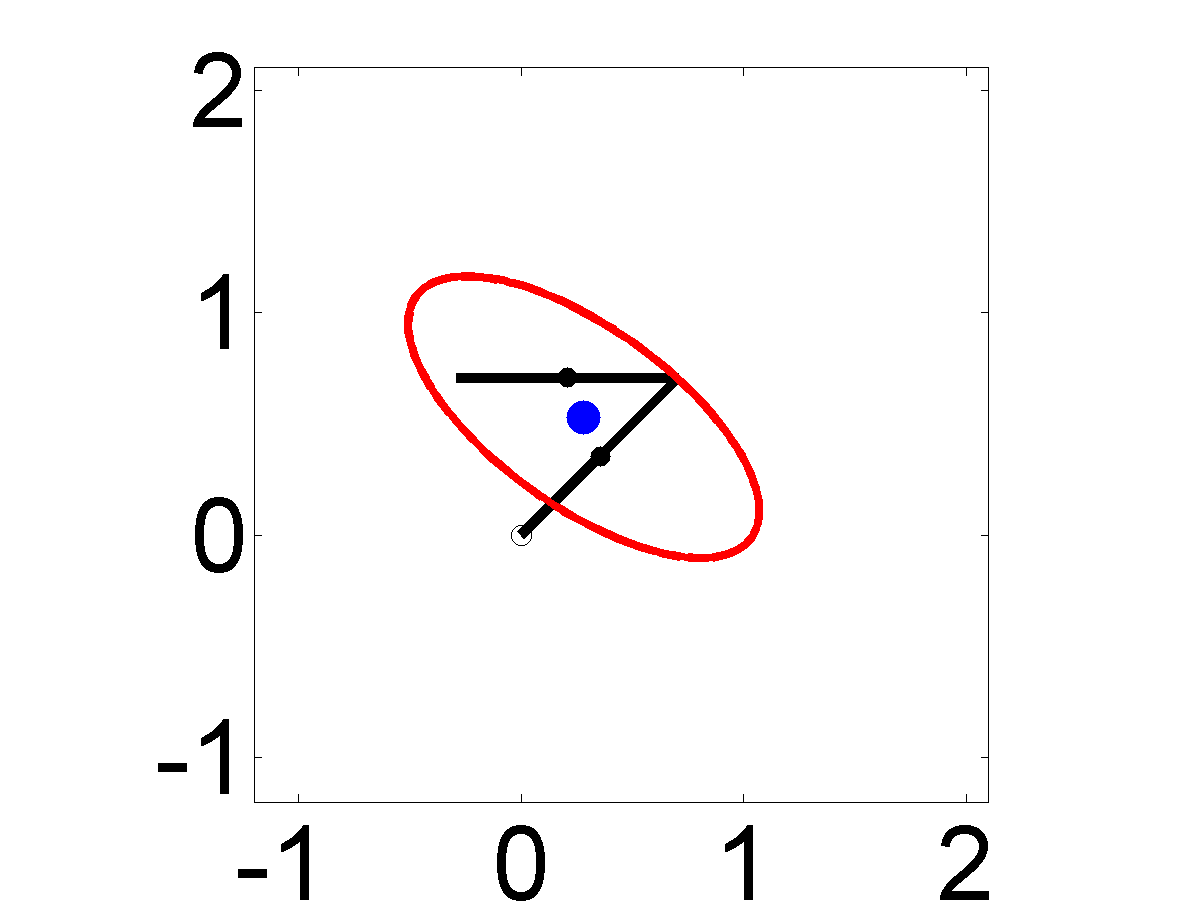}
\end{subfigure}
\begin{subfigure}{.115\textwidth}
  \centering
  \includegraphics[width=\linewidth]{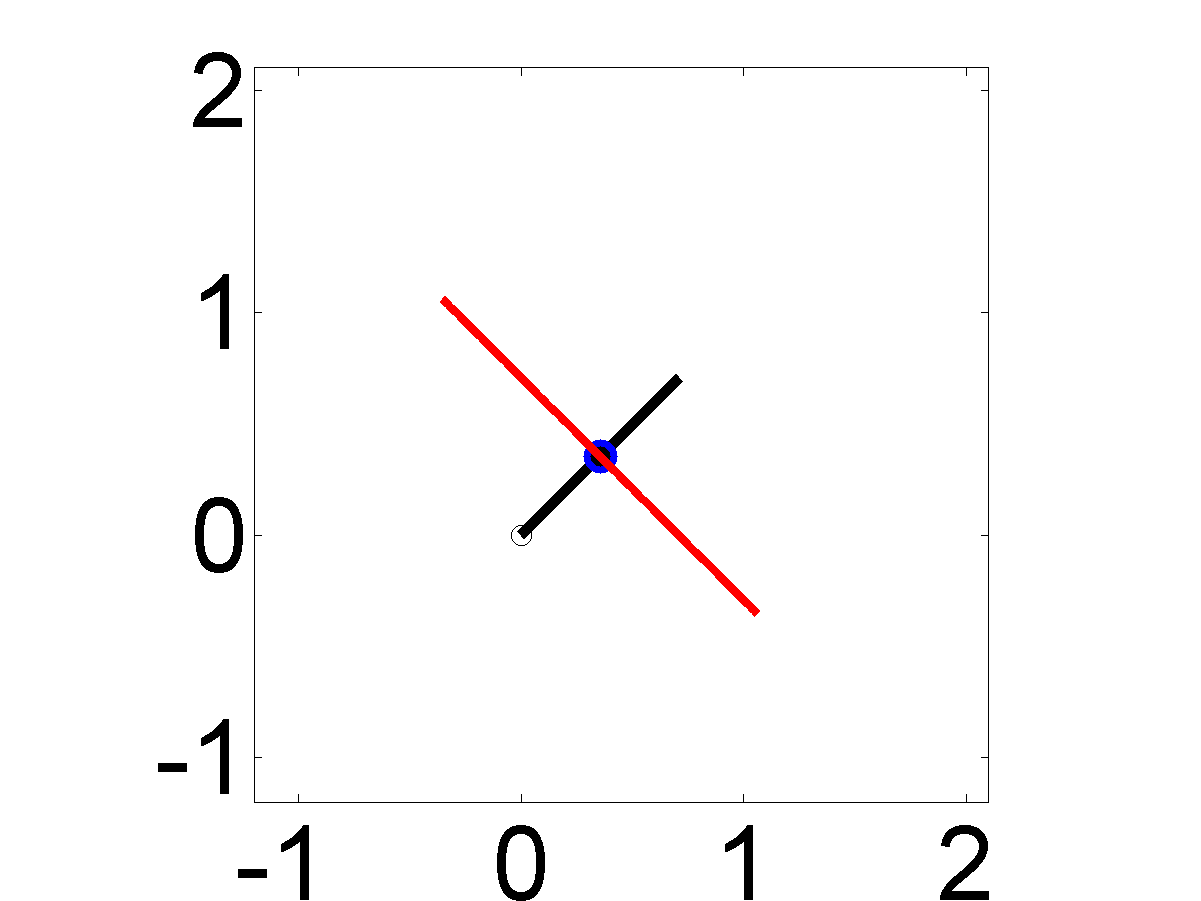}
\end{subfigure}
\begin{subfigure}{.115\textwidth}
  \centering
  \includegraphics[width=\linewidth]{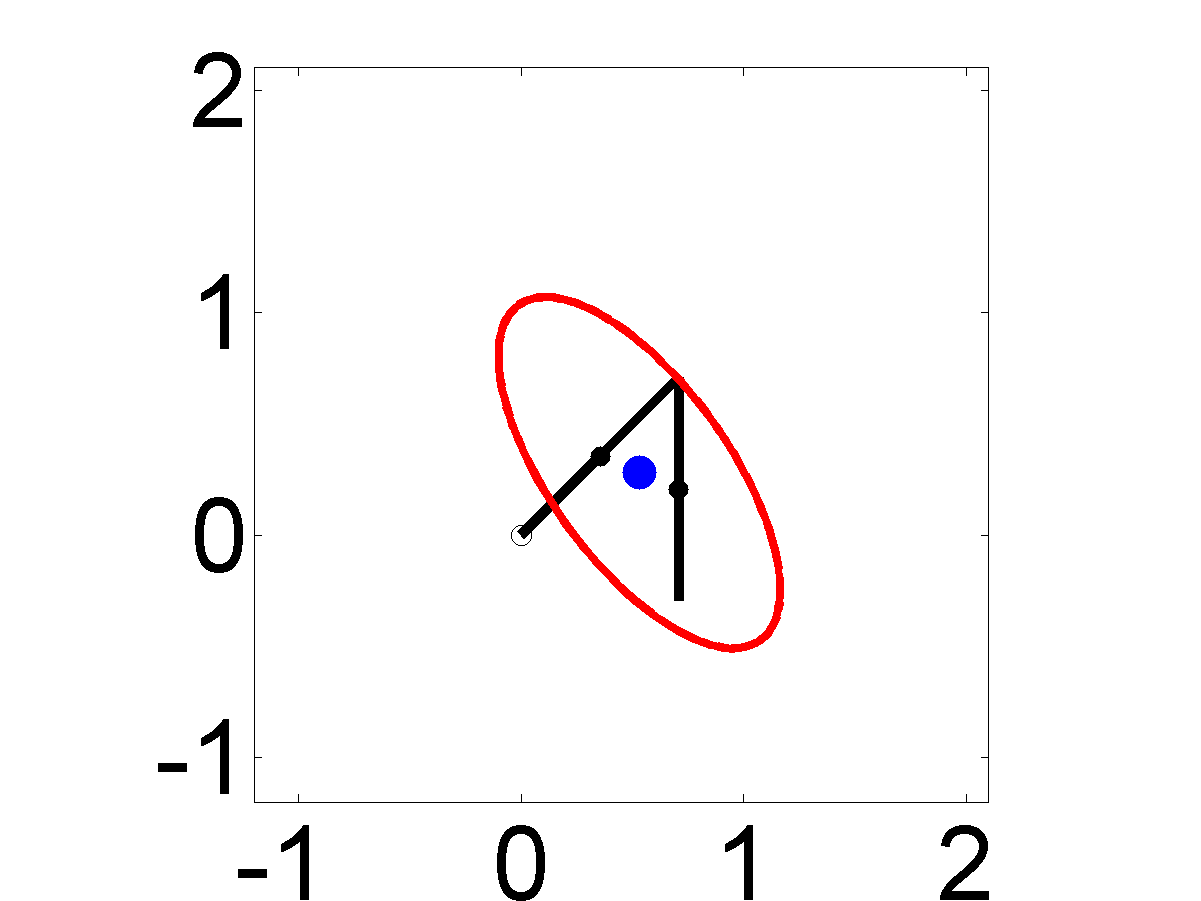}
\end{subfigure}
\begin{subfigure}{.115\textwidth}
  \centering
  \includegraphics[width=\linewidth]{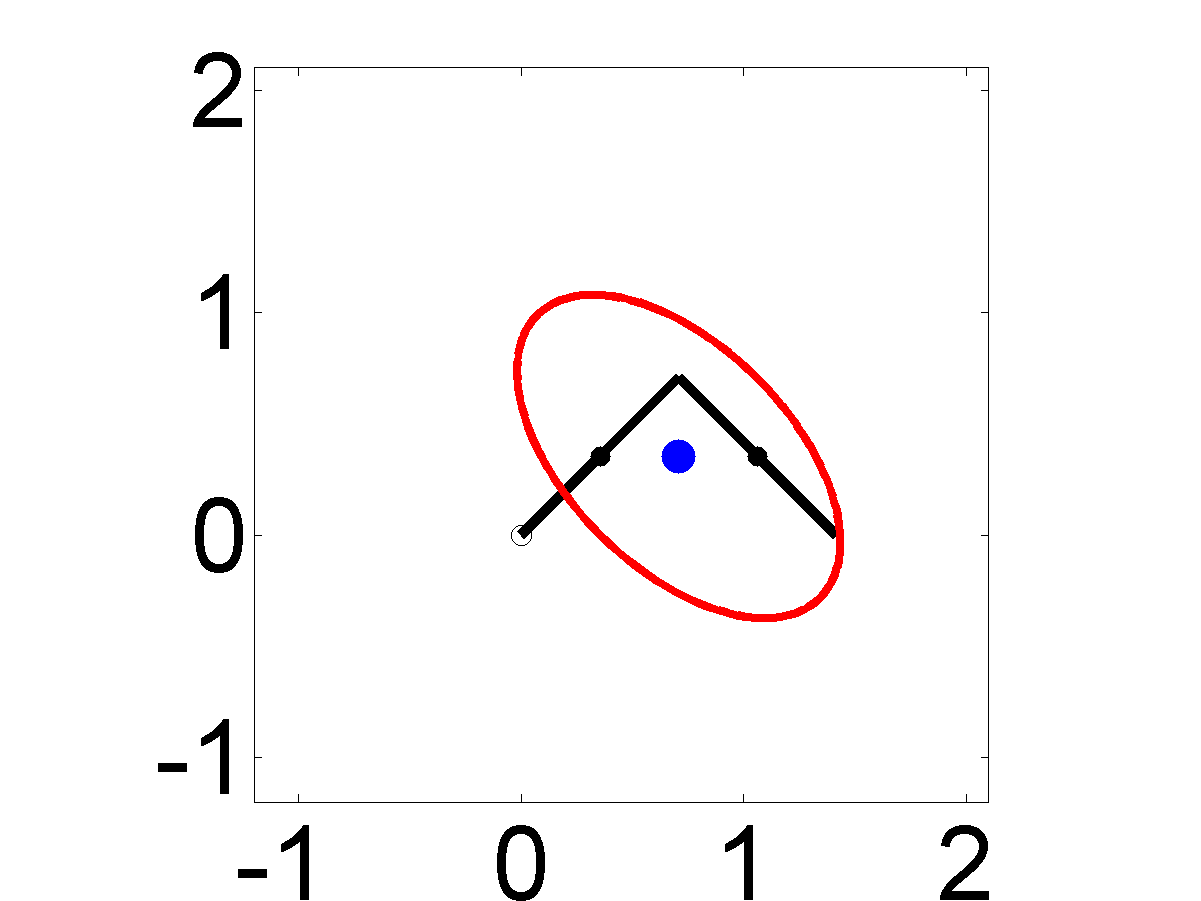}
\end{subfigure}
\begin{subfigure}{.115\textwidth}
  \centering
  \includegraphics[width=\linewidth]{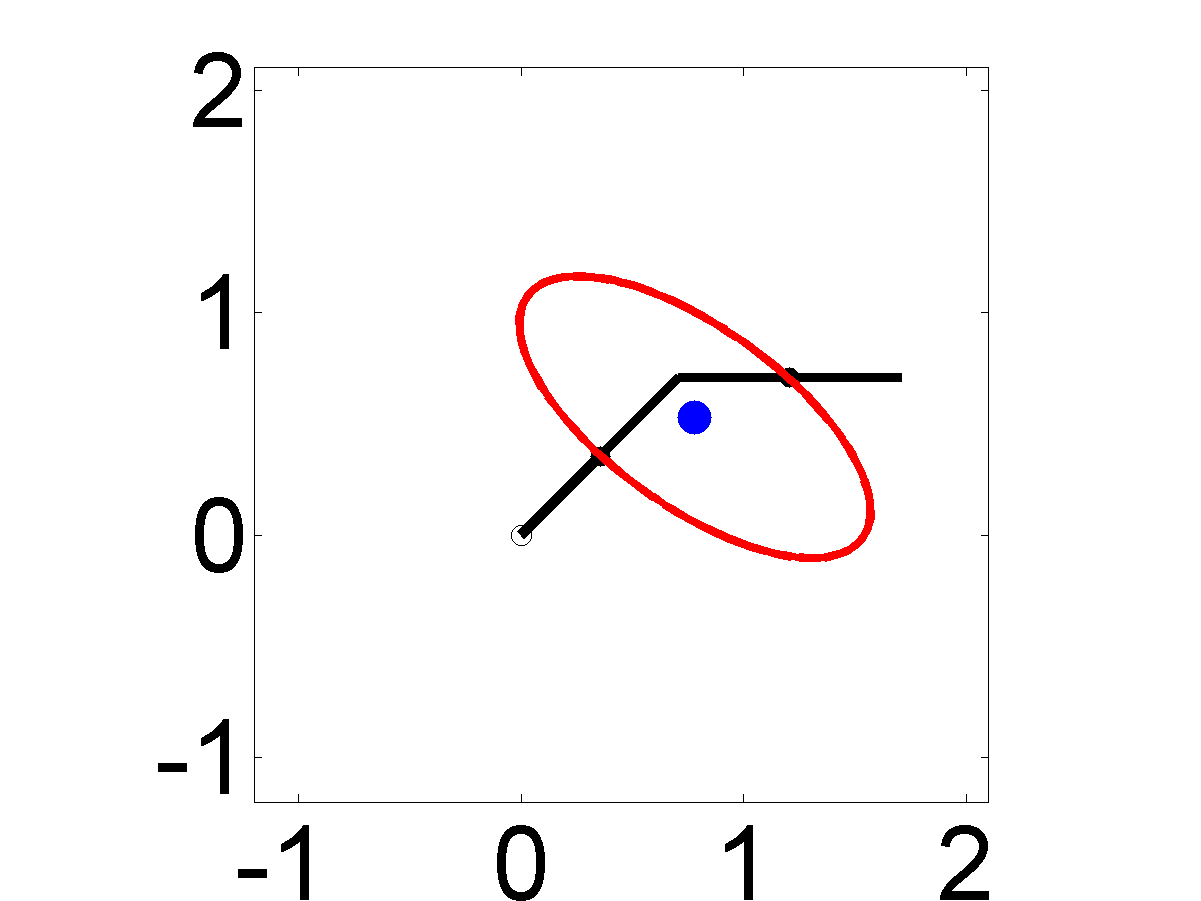}
\end{subfigure}
\begin{subfigure}{.115\textwidth}
  \centering
  \includegraphics[width=\linewidth]{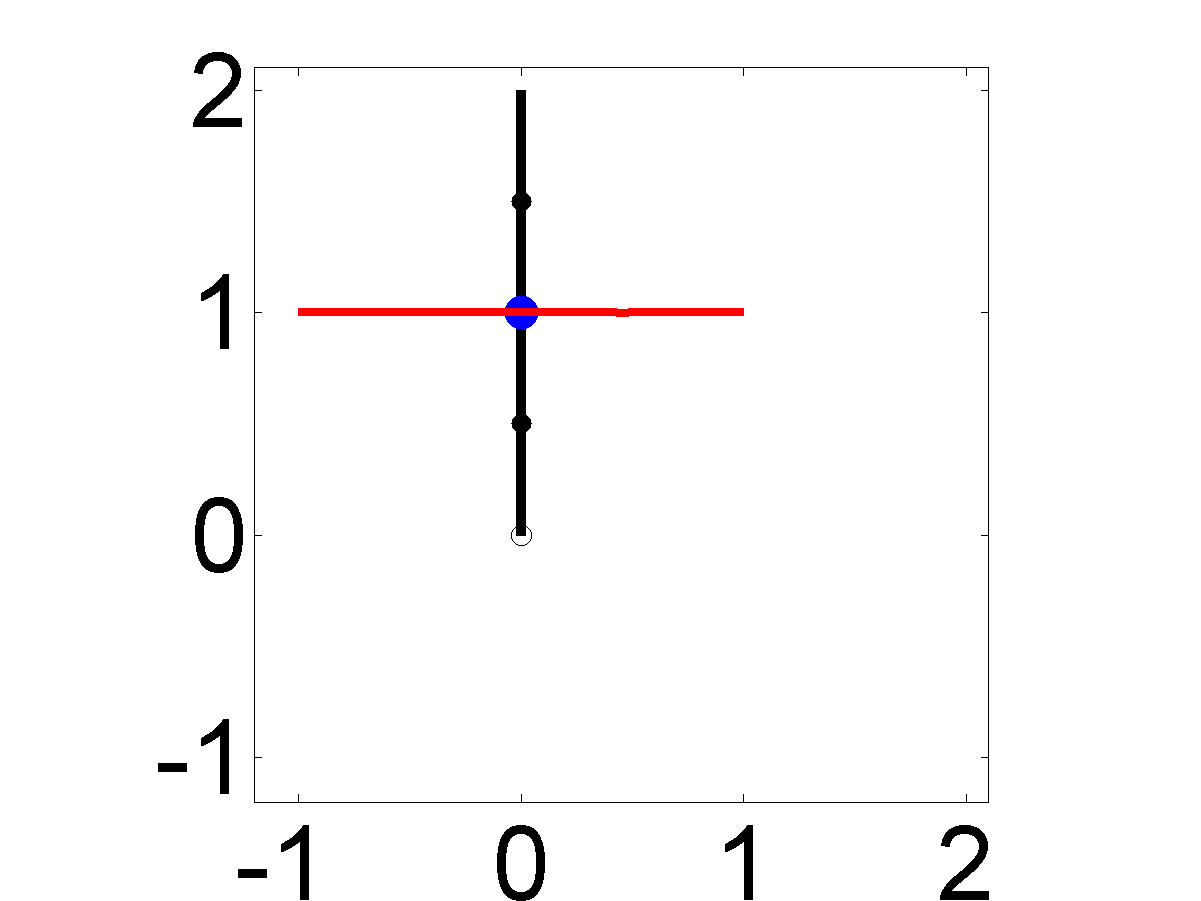}
\end{subfigure}
\begin{subfigure}{.115\textwidth}
  \centering
  \includegraphics[width=\linewidth]{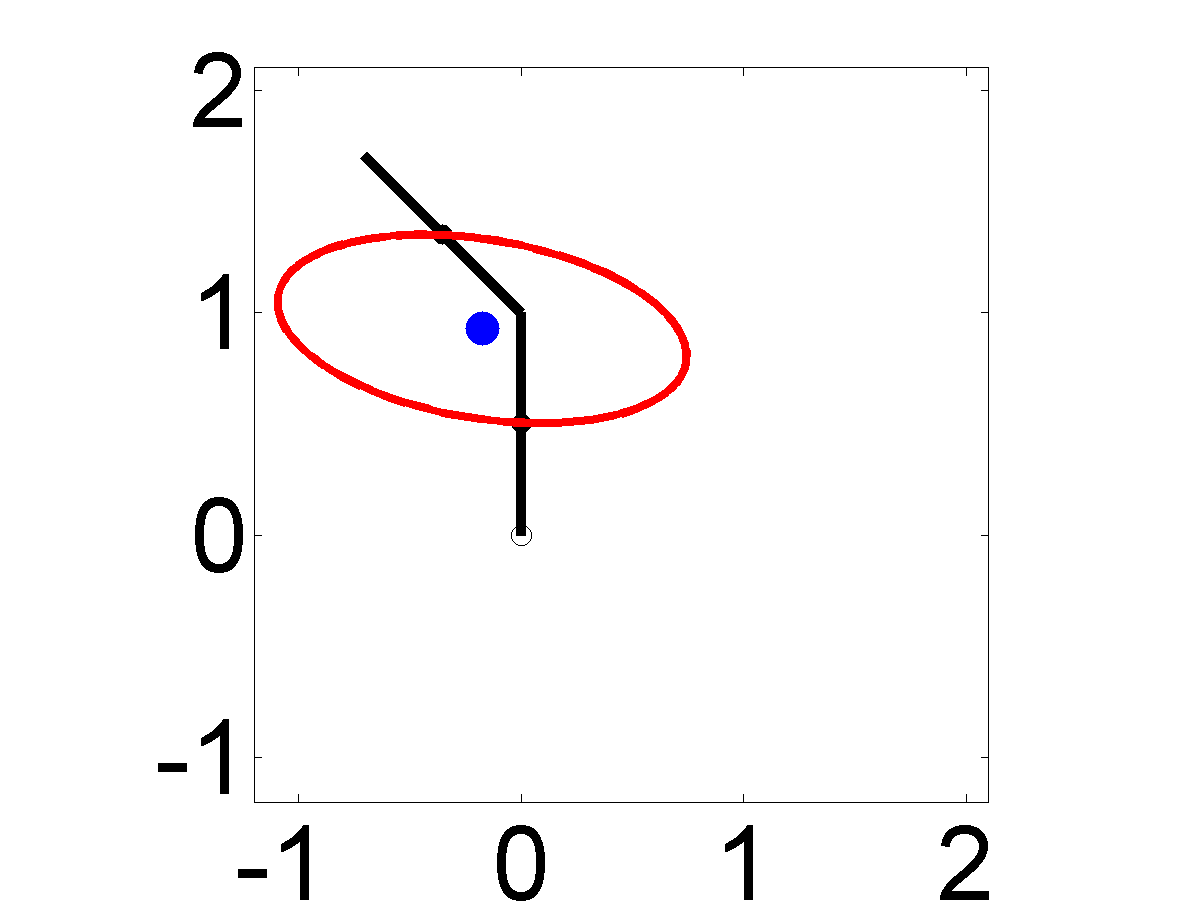}
\end{subfigure}
\begin{subfigure}{.115\textwidth}
  \centering
  \includegraphics[width=\linewidth]{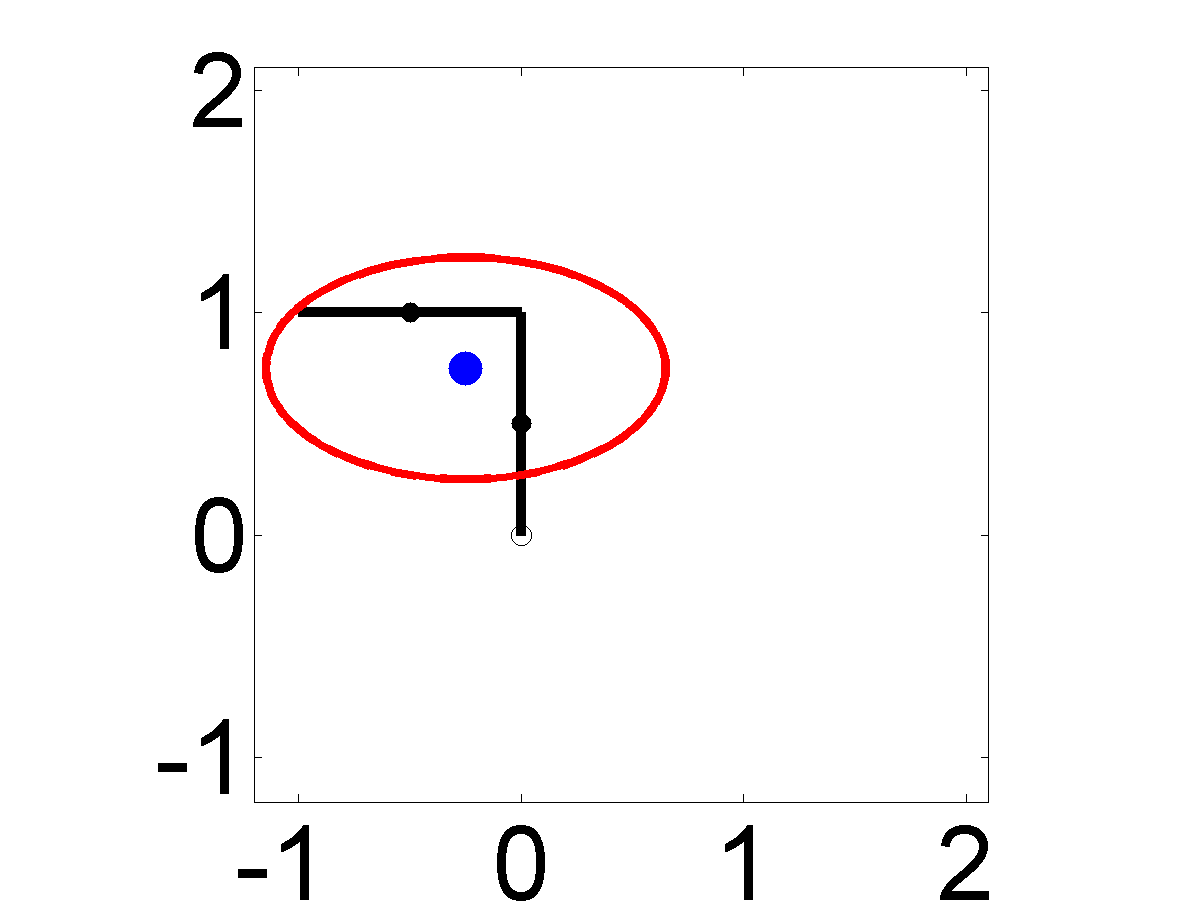}
\end{subfigure}
\begin{subfigure}{.115\textwidth}
  \centering
  \includegraphics[width=\linewidth]{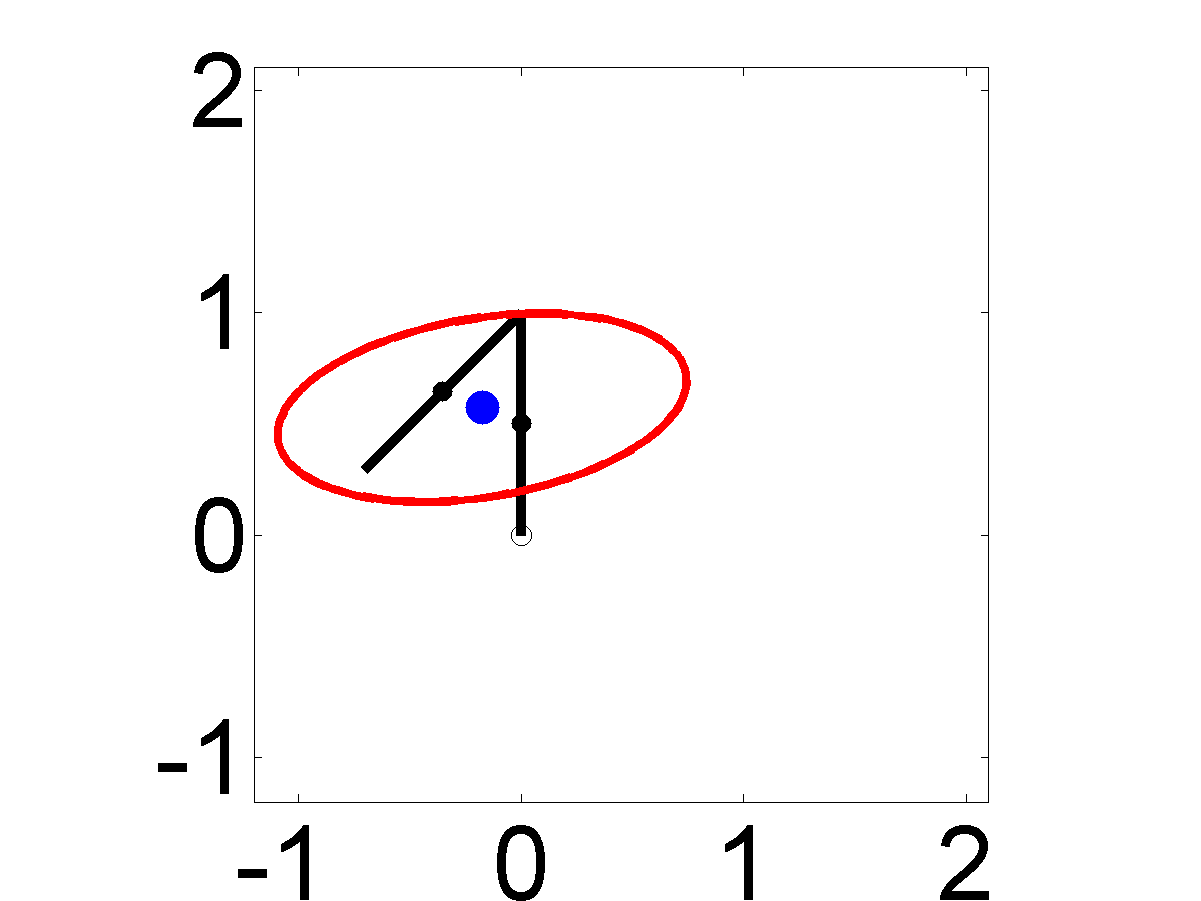}
\end{subfigure}
\begin{subfigure}{.115\textwidth}
  \centering
  \includegraphics[width=\linewidth]{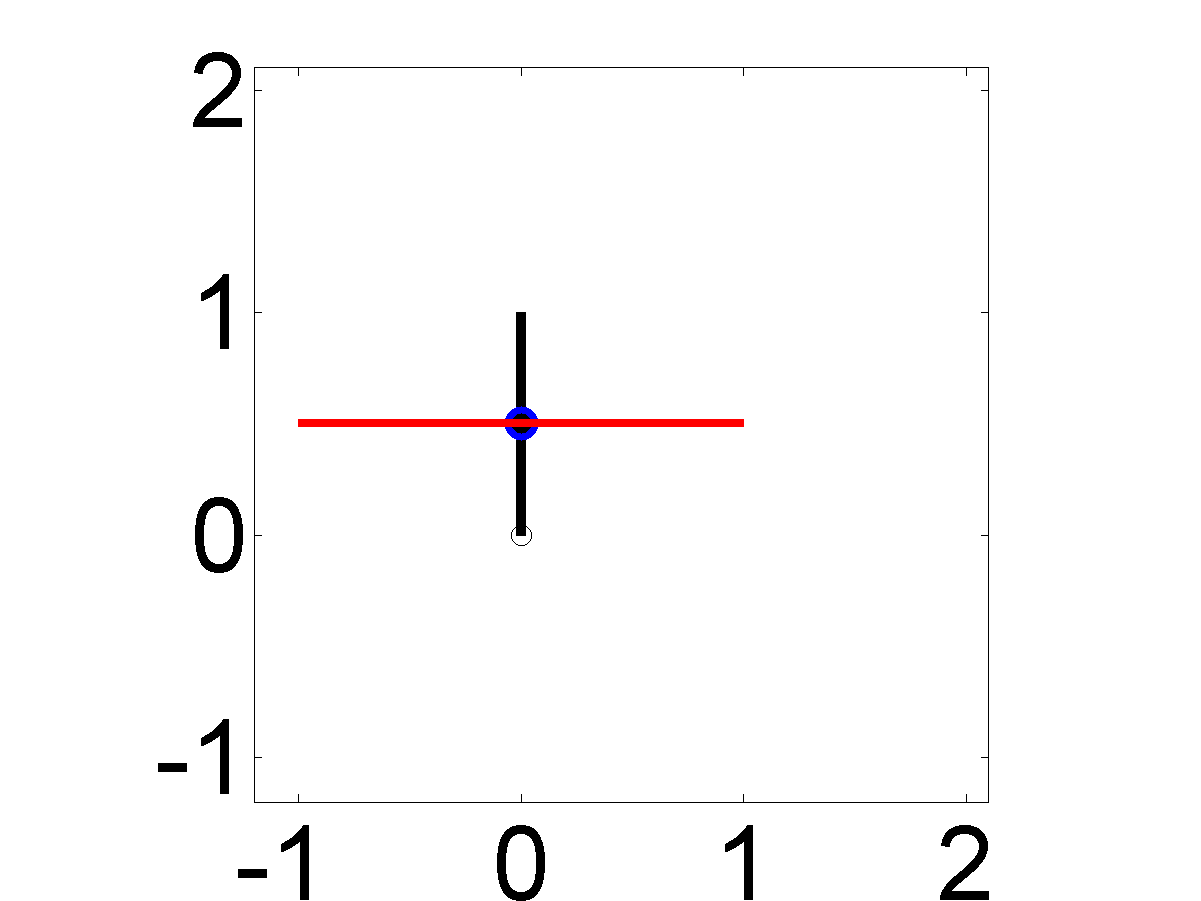}
\end{subfigure}
\begin{subfigure}{.115\textwidth}
  \centering
  \includegraphics[width=\linewidth]{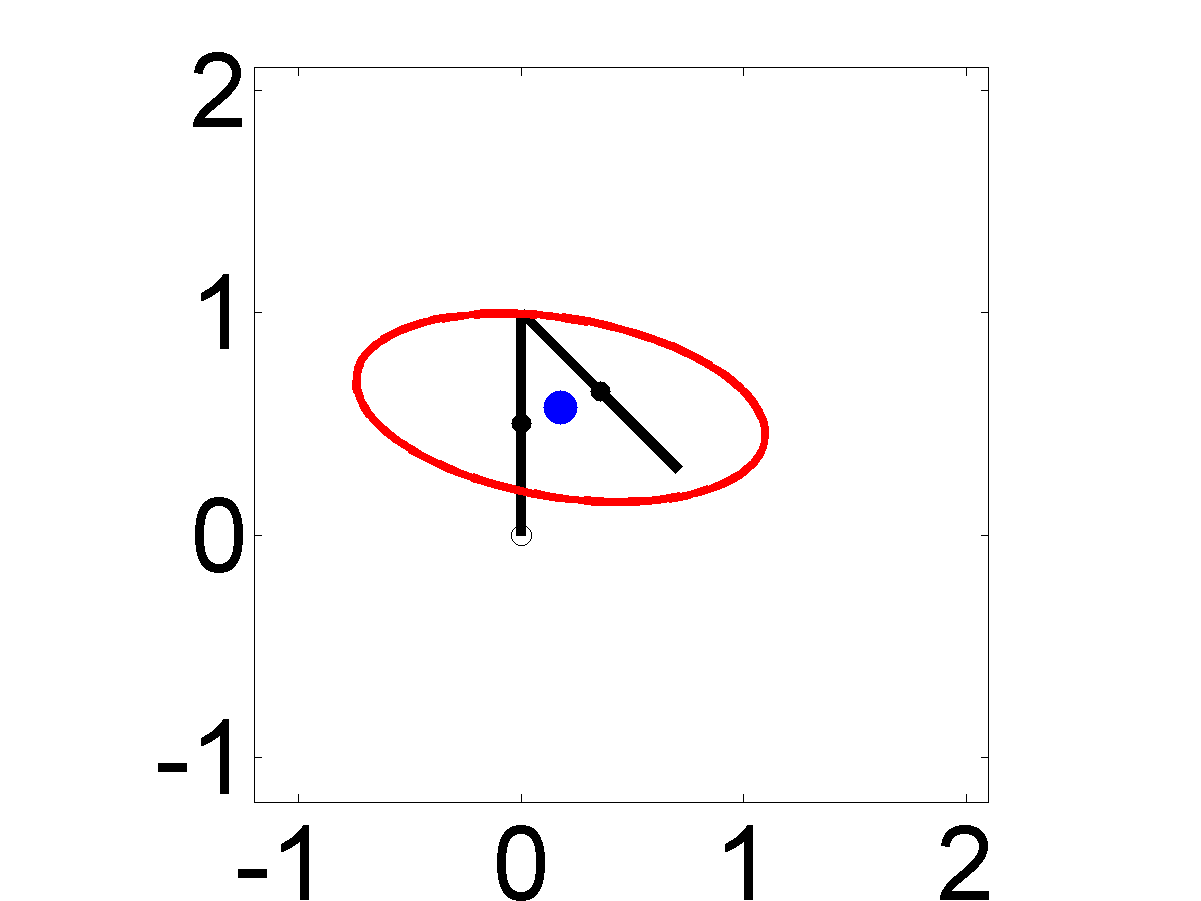}
\end{subfigure}
\begin{subfigure}{.115\textwidth}
  \centering
  \includegraphics[width=\linewidth]{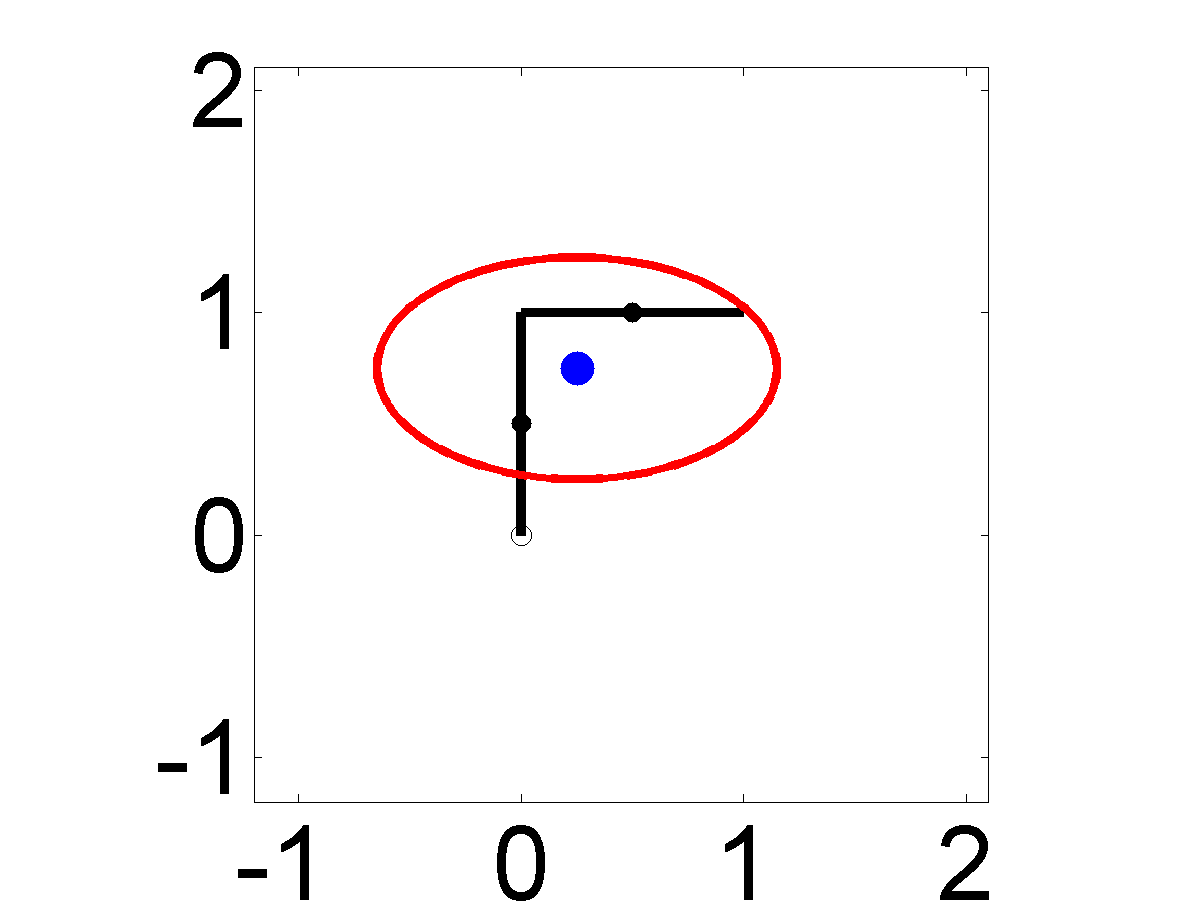}
\end{subfigure}
\begin{subfigure}{.115\textwidth}
  \centering
  \includegraphics[width=\linewidth]{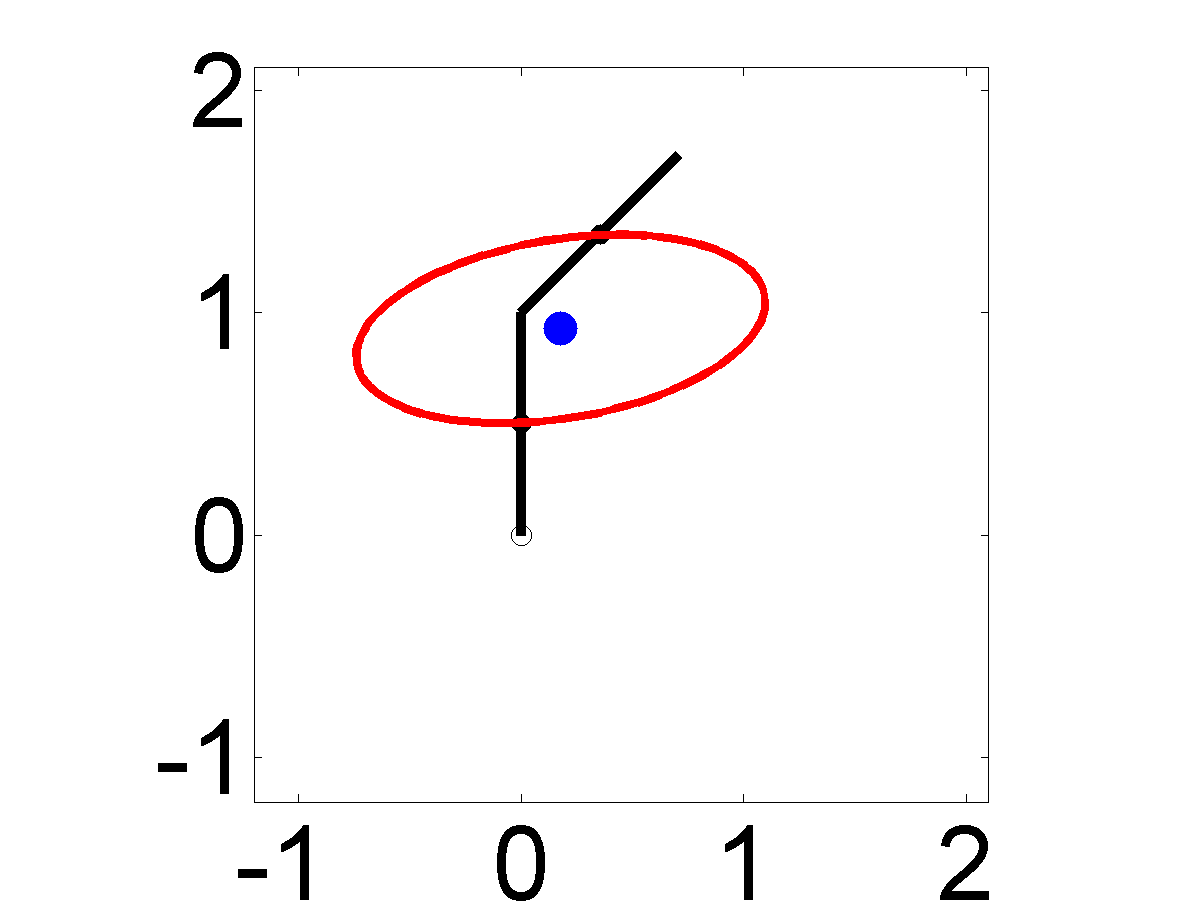}
\end{subfigure}

\caption{The transmissibility ellipsoid (ellipse in 2D) at different robot configurations in the case of a planar double pundulum}
\label{fig:ellipsoid}
\end{figure*}

Overall, the resultant of all active and passive forces acting of the system is $f$:

\begin{equation}
\begin{cases}
f_x = F_x \sin(q) \sin(q) - F_y \sin(q) \cos(q)\\
f_y = - F_x \sin(q) \cos(q) + F_y \cos(q) \cos(q)
\end{cases}
\end{equation}

It can be expressed in matrix form:

\begin{align}
\begin{split}
\begin{bmatrix} f_x \\ f_y \end{bmatrix}=\begin{bmatrix} \sin(q) \sin(q) & -\sin(q) \cos(q) \\ -\sin(q) \cos(q) & \cos(q) \cos(q) \end{bmatrix} \begin{bmatrix} F_x \\ F_y \end{bmatrix} =\\
=\begin{bmatrix} \sin(q) \sin(q) & -\sin(q) \cos(q) \\ -\sin(q) \cos(q) & \cos(q) \cos(q) \end{bmatrix} \begin{bmatrix} F \cos(\alpha) \\ F \sin(\alpha) \end{bmatrix}
\end{split}
\end{align}

Being $f$ the resultant of all forces, the dynamics of the system is described by the following relation:

\begin{equation}
\begin{cases}
\ddot x = \frac{1}{m} f_x\\
\ddot y = \frac{1}{m} f_y
\end{cases}
\end{equation}

Combining Eqs. 23 and 24 results in the overall description of the dynamic behavior of the pendulum:

\begin{align}
\begin{bmatrix} \ddot x \\ \ddot y \end{bmatrix}=\frac{1}{m} \begin{bmatrix} \sin(q) \sin(q) & -\sin(q) \cos(q) \\ -\sin(q) \cos(q) & \cos(q) \cos(q) \end{bmatrix} \begin{bmatrix} \cos(\alpha) \\ \sin(\alpha) \end{bmatrix} ||F||
\end{align}

It is now easy to identify the {\it Ratio of Transmission of Motion (RoToM)} of the pendulum, that depends on its configuration $q$, and on the direction $\alpha$ of the force vector $F$:
\begin{equation}
0 \le \begin{Vmatrix} \sin(q) \sin(q) \cos(\alpha) -\sin(q) \cos(q) \sin(\alpha)\\ -\sin(q) \cos(q) \cos(\alpha) +\cos(q) \cos(q) \sin(\alpha) \end{Vmatrix} \le 1
\end{equation}

The {\it RoToM} is always between 0 and 1. In particular, it will be equal to 0 if $\alpha = q + n \pi$, $\forall n \in \mathbb{Z}$:
\begin{align}
\begin{split}
\begin{Vmatrix} \sin(q) \sin(q) \cos(q) -\sin(q) \cos(q) \sin(q)\\ -\sin(q) \cos(q) \cos(q) +\cos(q) \cos(q) \sin(q) \end{Vmatrix} =\\
= \begin{Vmatrix} 0\\0 \end{Vmatrix} = 0
\end{split}
\end{align}

Instead, it will be equal to 1 if $\alpha = q + \frac{\pi}{2} + n \pi$, $\forall n \in \mathbb{Z}$:
\begin{align}
\begin{split}
\begin{Vmatrix} -\sin(q) \sin(q) \sin(\alpha) -\sin(q) \cos(q) \cos(q)\\ \sin(q) \cos(q) \sin(q) +\cos(q) \cos(q) \cos(q) \end{Vmatrix} =\\
= \begin{Vmatrix} -\sin(q)\\\cos(q) \end{Vmatrix} = 1
\end{split}
\end{align}

\begin{figure}[!t]
\centering
\begin{subfigure}{0.5\textwidth}
  \centering
  \includegraphics[width=\linewidth]{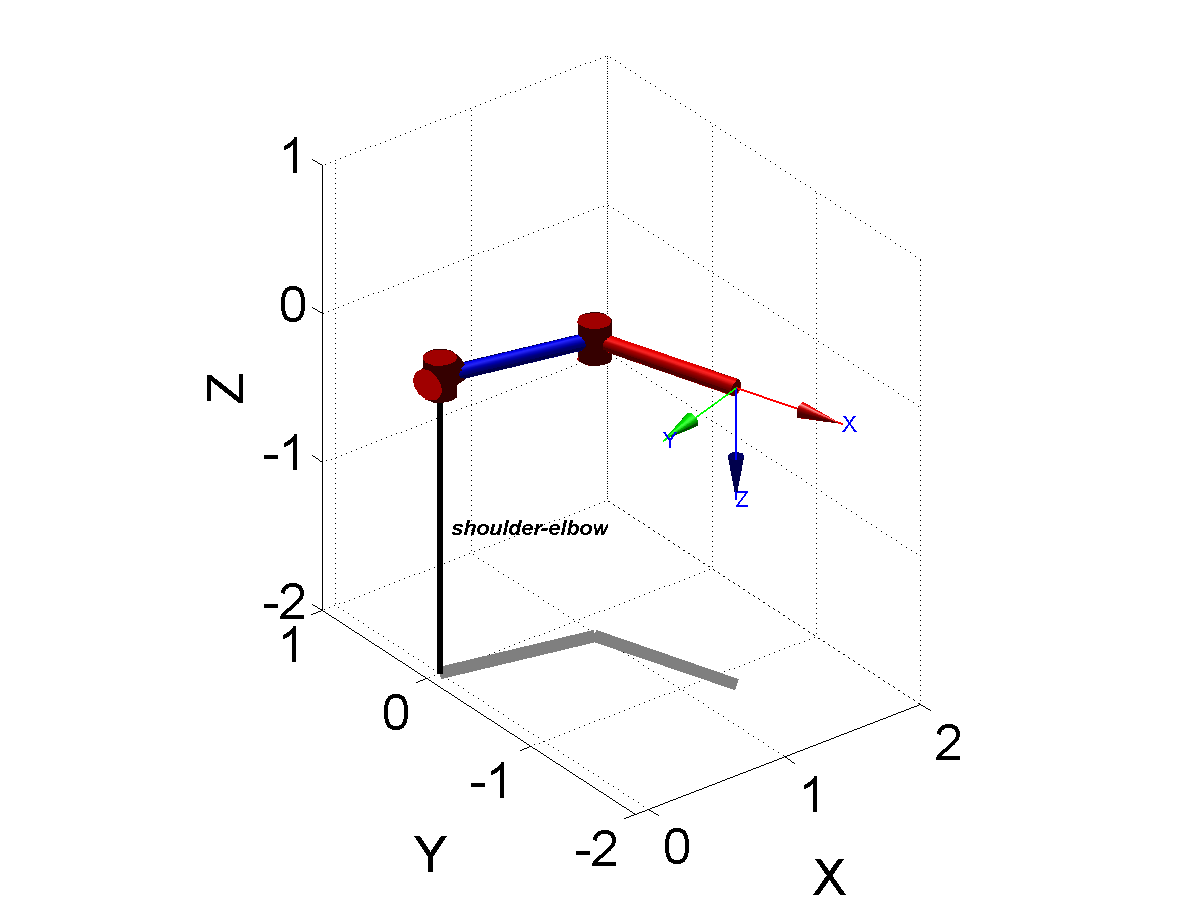}
\end{subfigure}
\begin{subfigure}{0.5\textwidth}
  \centering
  \includegraphics[width=\linewidth]{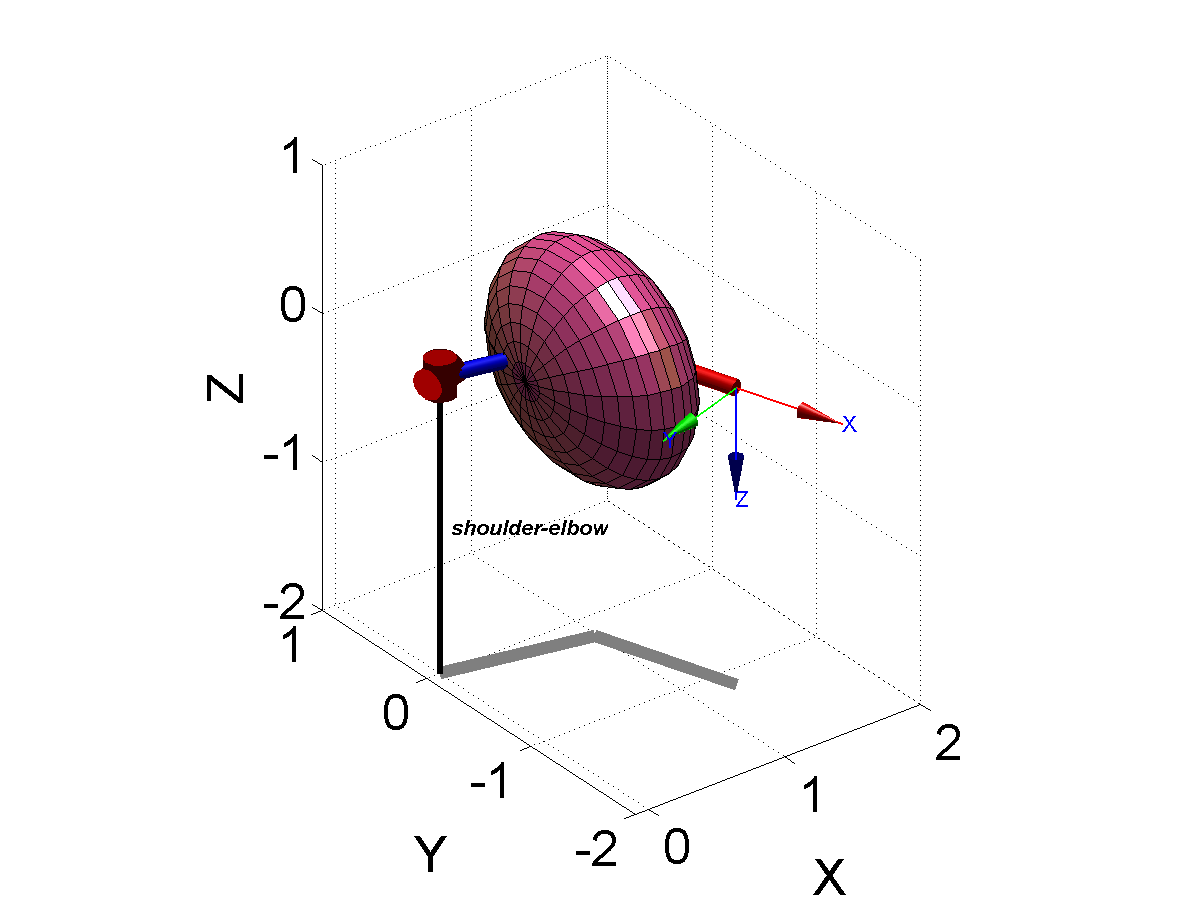}
\end{subfigure}
\caption{The transmissibility ellipsoid of a 4 degrees of freedom manipulator (shoulder-elbow) moving in 3D}
\label{fig:ellipsoid3d}
\end{figure}

\section{THE TRANSMISSIBILITY ELLIPSOID}
The \emph{Ratio of Transmission of Motion (RoToM)} is a measure of what part of a force that is applied to the CoM of a system contributes to its motion, and what part goes dissipated by the passive forces that are due to the mechanical constraints of the system itself. This value depends on the configuration of the robot $q$, and on the direction of the force vector $F$.

It could be useful to know what the dynamic behavior of the system will be when any force is applied to its CoM. This information is encoded in the \emph{transmissibility ellipsoid}, which is defined similarly to the manipulability ellipsoid \cite{Yoshikawa}. It is centered in the CoM of the robot $x_c$, and has semi-principal axes with direction and length defined by the eigenvectors $v$ and eigenvalues $\lambda$ of $m \Lambda^{-1}$, respectively. The transmissibility ellipsoid, hence, depends on the configuration $q$ of the robot (as shown in Figure \ref{fig:ellipsoid}; in Figure \ref{fig:ellipsoid3d} the transmissibility ellipsoid of a 4 degrees of freedom manipulator), and provides a visualization of what its dynamic behavior will be given the direction of an applied force.

\begin{figure}[!t]
\vspace{0.5cm}
\centering
\includegraphics[width=\linewidth]{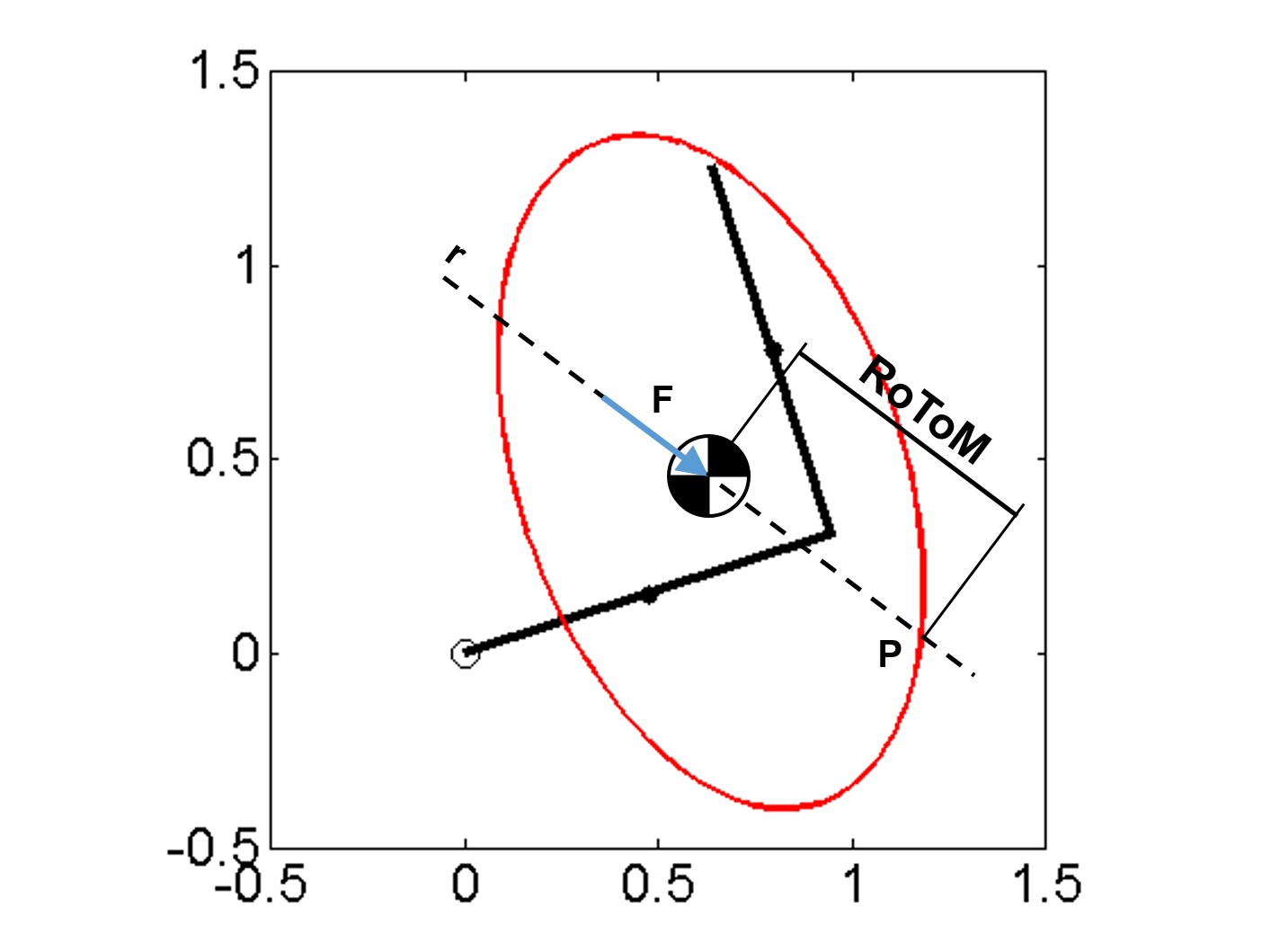}
\caption{The relation between the transmissibility ellipsoid (ellipse in 2D) and the \emph{RoToM} given a force vector $F$}\label{fig:ellipsoid_big}
\end{figure}

In fact, knowing the transmissibility ellipsoid and the direction of an external force that is applied to the CoM, it is possible to obtain the \emph{RoToM} (Figure \ref{fig:ellipsoid_big}):
\begin{itemize}
\item $r$ is the straight line with direction of the force vector, and passing by the center of the ellipsoid $x_c$
\item $P$ is the intersection (either of the two) between $r$ and the surface of the ellipsoid
\item the \emph{RoToM} is equal to the distance between the center of the ellipsoid $x_c$ and the point $P$
\end{itemize}

\begin{figure}[!t]
\vspace{0.5cm}
\centering
\includegraphics[width=\linewidth]{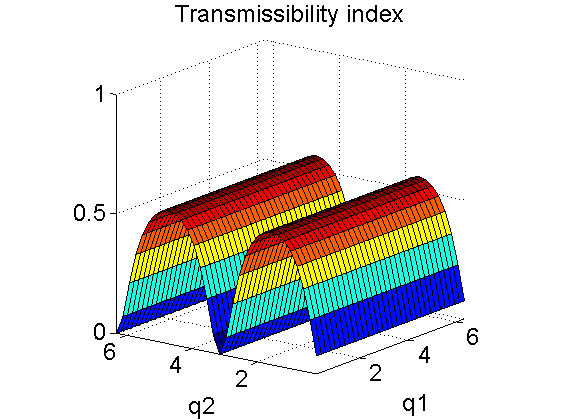}
\caption{The transmissibility index as a function of the robot configuration $q$ in the case of a planar double pendulum, with links with equal length and mass}\label{fig:Tindex}
\end{figure}

Another interesting measure that is related to the transmissibility ellipsoid is the \emph{transmissibility index}. It is defined as the ratio between the minor of the eigenvalues and the major of the eigenvalues of $m \Lambda^{-1}$. This value is always between 0 and 1, and provides an indication on how similarly the system behaves when subject to a force coming from different directions. If the transmissibility index is equal to 1, then the transmissibility ellipsoid is a sphere, and the \emph{RoToM} is the same for any direction of the applied force. On the contrary, if the transmissibility index is equal to 0, then the \emph{RoToM} is also equal to 0 in the direction of the semi-minor axis of the ellipsoid. The transmissibility index does not depend on the orientation of the robot with respect to the world, but on the reciprocal orientation of the links (in Figure \ref{fig:Tindex} the transmissibility index does not depend on the angle of the first joint, but on the angle of the second joint only).

All this information is particularly important in the case of redundant robots, when several configurations are valid for achieving the same task(s). In this case it is possible to chose a certain configuration among the others, in order to have e.g., a transmissibility ellipsoid that is as close as possible to a desired one. The reference ellipsoid will depend on the application and on the desired behavior of the system. In the instance of gravity only acting on the system it could be beneficial to have an ellipsoid with semi-minor axis that is as vertical as possible, and with minimal length. Conversely, if the robot is about to be subject to an external disturbance, but its direction is not known in advance, it could be good to have a transmissibility ellipsoid that is as close as possible to a sphere (i.e., with a transmissibility index that is equal to one).

\section{POSSIBLE GENERALIZATION}
What if one or more external forces are applied to the robot at different locations? In the most general case the resultant of all active forces at the CoM is a wrench, i.e., it has both linear and rotational components. This means that the system cannot be reduced to a point mass with mass $m$ as in the previous case. An equivalent inertia $I$ should also be considered. Differently from $m$, though, $I$ is not a constant, but is a function of $q$, instead. Moreover, since $I$ can vary, when a centroidal torque is applied there is no maximal rotational acceleration. Ideally the robot can move to configurations such that the inertia in that specific direction is equal to zero, resulting in an infinite acceleration.

For this very last reason it is not possible to define a {\it Ratio of Transmission of Motion (RoToM)} for the rotation. It is nevertheless still possible to modify the configuration of the robot in order to locally maximize or minimize the effects of a certain torque that is applied in terms of rotational centroidal acceleration.

A second consideration is that, given a wrench that is applied to the CoM, there might be changes in the configuration that maximize/minimize the linear acceleration but that are in conflict with changes in the configuration that maximize/minimize the rotational acceleration. Being linear and rotational acceleration different quantities it is not straightforward to define a unique scalar indicator to be maximized/minimized (e.g., the norm of both linear and rotational components \cite{Ferrari}, \cite{Schulman}). It is still possible to have a linear combination of the norm of the linear component and of the norm of the rotational components. This will also make it possible to favor the minimization/maximization of one over the other just by adjusting a single coefficient.

\section{CONCLUSIONS}
This paper presents an analysis of the dynamic response of a robot when subject to an external force that is applied to its Center of Mass (CoM). The \emph{Ratio of Transmission of Motion (RoToM)} is proposed as a novel indicator of what part of the applied force generates motion, and what part is dissipated by the passive forces due to mechanical constraints. The \emph{RoToM} is a scalar quantity between 0 and 1 that depends on the configuration of the robot, and on the direction of the applied force. It is instead independent of both mass of the robot, and magnitude of the applied force.

Depending on the application it may be beneficial to have a high \emph{RoToM} or a low \emph{RoToM}. In the case, for instance, of a robot that is subject to gravity only, a smaller \emph{RoToM} requires a lesser effort in terms of joint torques to compensate for it. A local minimization using a gradient descent can be very useful to solve practical problems of this kind. However, in the case of redundant robots that typically have several \emph{RoToM} minima, it is always possible to identify all the configurations with zero \emph{RoToM}.

Extending the concept of \emph{RoToM}, the \emph{transmissibility ellipsoid} is presented. It depends on the configuration of the robot, and provides a visualization of what its dynamic behavior will be given the direction of an external force that is applied to its CoM.

Another physical measure that is related to the transmissibility ellipsoid is the \emph{transmissibility index}: it is a scalar value between 0 and 1, and provides an indication on how similarly the system behaves when subject to forces coming from different directions. The transmissibility index does not depend on the orientation of the robot with respect to the world, but on the reciprocal orientation of the links.

The presented analysis aims to provide a deeper insight on the centroidal dynamics of a robot. It shows the importance of the role played by the centroidal inertia, a quantity that depends on the robot configuration, in the dynamic behavior of the robot.

Possible applications include the development of whole-body controllers, for e.g., minimizing the overall effort of the system in terms of joint torques due to gravity, and the design of interaction control architectures.

Further studies will be performed to extend the presented concepts to rotational quantities, as described in Section V. Developing a formulation that accounts for both linear and rotational applied forces will make the analysis more general, as it will be suitable to describe the dynamic behavior of a robot that is subject to any force with an arbitrary application point.

\addtolength{\textheight}{-12cm}   % This command serves to balance the column lengths

\section*{ACKNOWLEDGMENT}
Federico L. Moro warmly thanks Michael Gienger, Honda RI, and Niccol{\`o} Iannacci, ITIA-CNR for the helpful discussions on the topics presented in this paper.

\end{document}